\documentclass[english,twocolumn]{svjour3}
\usepackage[T1]{fontenc}
\usepackage[latin9]{inputenc}
\usepackage{xcolor}
\usepackage{array}
\usepackage{calc}
\usepackage{url}
\usepackage{multirow}
\usepackage{amsmath}
\usepackage{amssymb}
\usepackage{graphicx}
\usepackage{microtype}

\makeatletter

\providecommand{\tabularnewline}{\\}

\usepackage[T1]{fontenc}
\usepackage{lmodern}

\usepackage{times}
\usepackage{epsfig}
\usepackage{graphicx}
\usepackage{microtype}
\usepackage{balance}

\usepackage[linesnumbered,algoruled,boxed,lined]{algorithm2e}
\usepackage{siunitx}
\usepackage{amssymb,amsmath}

\makeatother

\usepackage{babel}
\begin{document}
\title{Hierarchical Conditional Relation Networks for Multimodal Video Question
Answering}
\titlerunning{HCRN for Multimodal Video QA }
\author{Thao Minh Le \and  Vuong Le \and  Svetha Venkatesh \and  Truyen
Tran}
\authorrunning{Le et al.}
\institute{Thao Minh Le, Vuong Le, Svetha Venkatesh and Truyen Tran \at  Applied
AI Institute, Deakin University, Australia\\
\email{lethao@deakin.edu.au}\\
}
\date{Received: date / Accepted: date}

\maketitle
\global\long\def\Problem{\text{Video QA}}%

\global\long\def\ModelName{\text{HCRN}}%
\global\long\def\UnitName{\text{CRN}}%

\begin{abstract}
Video Question Answering (Video QA) challenges modelers in multiple
fronts. Modeling video necessitates building not only spatio-temporal
models for the dynamic visual channel but also multimodal structures
for associated information channels such as subtitles or audio. Video
QA adds at least two more layers of complexity -- selecting relevant
content for each channel in the context of the linguistic query, and
composing spatio-temporal concepts and relations hidden in the data
in response to the query. To address these requirements, we start
with two insights: (a) content selection and relation construction
can be jointly encapsulated into a conditional computational structure,
and (b) video-length structures can be composed hierarchically. For
(a) this paper introduces a general-reusable reusable neural unit
dubbed Conditional Relation Network (CRN) taking as input a set of
tensorial objects and translating into a new set of objects that encode
relations of the inputs. The generic design of CRN helps ease the
common complex model building process of Video QA by simple block
stacking and rearrangements with flexibility in accommodating diverse
input modalities and conditioning features across both visual and
linguistic domains. As a result, we realize insight (b) by introducing
Hierarchical Conditional Relation Networks (HCRN) for Video QA. The
HCRN primarily aims at exploiting intrinsic properties of the visual
content of a video as well as its accompanying channels in terms of
compositionality, hierarchy, and near-term and far-term relation.
HCRN is then applied for Video QA in two forms, short-form where answers
are reasoned solely from the visual content of a video, and long-form
where an additional associated information channel, such as movie
subtitles, presented. Our rigorous evaluations show consistent improvements
over state-of-the-art methods on well-studied benchmarks including
large-scale real-world datasets such as TGIF-QA and TVQA, demonstrating
the strong capabilities of our CRN unit and the HCRN for complex domains
such as Video QA. To the best of our knowledge, the HCRN is the very
first method attempting to handle long and short-form multimodal Video
QA at the same time.
\end{abstract}

\section{Introduction}

Answering natural questions about a video is a powerful demonstration
of cognitive capability. The task involves acquisition and manipulation
of spatio-temporal visual, acoustic and linguistic representations
from the video guided by the compositional semantics of linguistic
cues \cite{gao2018motion,lei2018tvqa,li2019beyond,song2018explore,tapaswi2016movieqa,wang2018movie}.
As questions are potentially unconstrained, $\Problem$ requires deep
modeling capacity to encode and represent crucial multimodal video
properties such as linguistic content, object permanence, motion profiles,
prolonged actions, and varying-length temporal relations in a hierarchical
manner. For $\Problem$, the visual and textual representations should
ideally be question-specific and answer-ready. 

\begin{figure*}
\begin{minipage}[c][1\totalheight][b]{0.48\textwidth}%
\begin{center}
\includegraphics[width=1\textwidth]{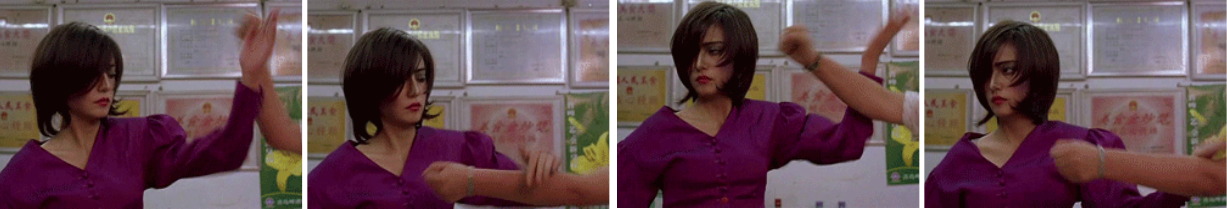}
\par\end{center}
(1) \textbf{Question}: What does the girl do 9 times?

\qquad{}\textbf{Choice 1}: walk

\qquad{}\textbf{Choice 2}: blocks a person's punch

\qquad{}\textbf{Choice 3}: step

\qquad{}\textbf{Choice 4}: shuffle feet

\qquad{}\textbf{Choice 5}: wag tail\medskip{}

\hspace{1.5em}Baseline: \textcolor{red}{walk}

\hspace{1.5em}$\ModelName$: \textcolor{green}{blocks a person's
punch}

\hspace{1.5em}Ground truth: \textcolor{brown}{blocks a person's punch}%
\end{minipage}\hfill{}%
\begin{minipage}[c][1\totalheight][b]{0.48\textwidth}%
\begin{center}
\includegraphics[width=1\textwidth]{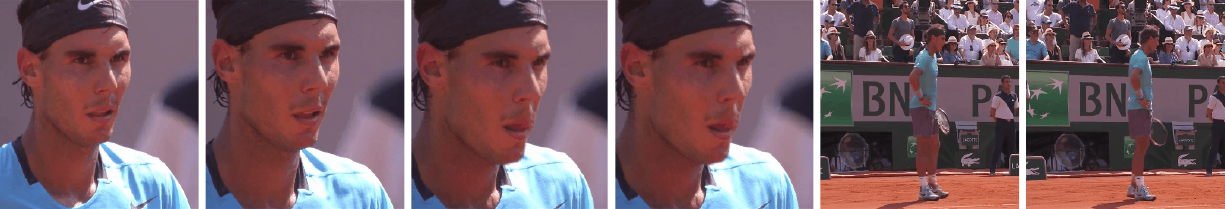}
\par\end{center}
(2) \textbf{Question}: What does the man do before turning body to
left?

\qquad{}\textbf{Choice 1}: run across a ring

\qquad{}\textbf{Choice 2}: pick up the man's hand

\qquad{}\textbf{Choice 3}: flip cover face with hand

\qquad{}\textbf{Choice 4}: raise hand

\qquad{}\textbf{Choice 5}: breath\medskip{}

\hspace{1.5em}Baseline: \textcolor{red}{pick up the man's hand}

\hspace{1.5em}$\ModelName$: \textcolor{green}{breath}

\hspace{1.5em}Ground truth: \textcolor{brown}{breath}%
\end{minipage}

\caption{Examples of \emph{short-form $\protect\Problem$} for which frame
relations are key toward correct answers. {\small{}(1) }\emph{\small{}Near-term
frame relations }{\small{}are required for counting of fast actions.
(2) }\emph{\small{}Far-term frame relations }{\small{}connect the
actions in long transition.} $\protect\ModelName$ with the ability
to model hierarchical conditional relations handles successfully,
while baseline struggles.\label{fig:Examples-of-short-form}}
\end{figure*}

\begin{figure*}
\noindent\begin{minipage}[t]{1\textwidth}%
\begin{center}
\includegraphics[width=0.98\textwidth]{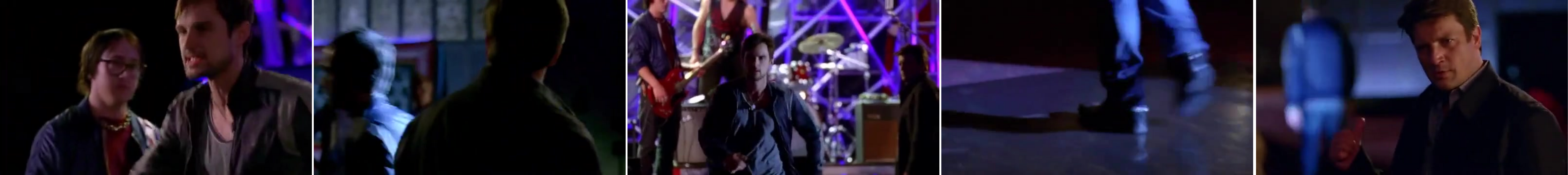}
\par\end{center}%
\end{minipage}

\medskip{}
\textbf{Subtitle}:\medskip{}

\noindent{\fboxrule 0.7pt\fboxsep 4pt\fbox{\begin{minipage}[c]{1\textwidth - 2\fboxsep - 2\fboxrule}%
\begin{center}
\begin{minipage}[t]{0.34\textwidth}%
\textcolor{teal}{00:00:0.395 -{}-> 00:00:1.896}

(Keith:) I'm not gonna stand here and let you accuse me%
\end{minipage}\hfill{}%
\begin{minipage}[t]{0.34\textwidth}%
\textcolor{teal}{00:00:1.897 -{}-> 00:00:4.210}

(Keith:) of killing one of my best friends, all right?%
\end{minipage}\hfill{}%
\begin{minipage}[t]{0.28\textwidth}%
\textcolor{teal}{00:00:8.851 -{}-> 00:00:10.394}

(Castle:) You hear that sound?%
\end{minipage}
\par\end{center}%
\end{minipage}}}
\begin{centering}
\medskip{}
\par\end{centering}
\noindent\begin{minipage}[t]{1\textwidth}%
\textbf{Question}: What did Keith do when he was on the stage?

\begin{tabular}{>{\raggedright}p{0.4\textwidth}>{\raggedright}p{0.5\textwidth}}
\quad{}\textbf{Choice 1}: Keith drank beer

\quad{}\textbf{Choice 2}: Keith played drum

\quad{}\textbf{Choice 3}: Keith sing to the microphone & \quad{}

\quad{}\textbf{Choice 4}: Keith played guitar

\quad{}\textbf{Choice 5}: Keith got off the stage and walked out\tabularnewline
\end{tabular}

\medskip{}

\hspace{1.5em}Baseline: \textcolor{red}{Keith played guitar}

\hspace{1.5em}$\ModelName$: \textcolor{green}{Keith got off the
stage and walked out}

\hspace{1.5em}Ground truth: \textcolor{brown}{Keith got off the stage
and walked out}%
\end{minipage}

\caption{Example of \emph{long-form} $\protect\Problem$. This is a typical
question where a model needs to collect sufficient relevant cues from
both visual content of a given video and textual subtitles to give
the correct answer. In this particular example, our baseline is likely
to suffer from the linguistic bias \emph{(``stage'' }and\emph{ ``played
guitar''}) while our model successfully manages to arrive at the
correct answer by connecting the linguistic information from the first
shot and visual content in the second one.\label{fig:Examples-of-long-form}}
\end{figure*}

The common approach toward modeling videos for QA is to build neural
architectures specially designed for a particular data format and
modality. In this perspective, the two common variants of $\Problem$
emerge: \emph{short-form $\Problem$} where relevant information is
contained in the visual content of short video snippets of a single
event (see Fig.~\ref{fig:Examples-of-short-form} for examples),
and \emph{long-form $\Problem$} (also called Video Story QA) where
the keys to answer are carried in the mixed visual-textual data of
longer multi-shot video sequences (see Fig.~\ref{fig:Examples-of-long-form}
for example). Because of this specificity, such hand crafted architectures
tend to be non-optimal for changes in data modality \cite{lei2018tvqa},
varying video length \cite{na2017read} or question types (such as
frame QA \cite{li2019beyond} versus action count \cite{fan2019heterogeneous}).
This has resulted in proliferation of heterogeneous networks.

We wish to build a family of effective models for both long-form and
short-form Video QA from reusable units that are more homogeneous
and easier to construct, maintain and comprehend. We start by realizing
that Video QA involves two sub-tasks: (a) selecting relevant content
for each channel in the context of the linguistic query, and (b) composing
spatio-temporal concepts and relations hidden in the data in response
to the query. Much of sub-task (a) can be abstracted into a conditional
computational structure that computes multi-way interaction between
the query and the several objects. With this ability, solving sub-task
(b) can be approached by composing the hierarchical structure of such
abstraction from the ground up. 

Toward this goal, we propose a general-purpose \emph{reusable neural
unit} called Conditional Relation Network ($\UnitName)$ that encapsulates
and transforms an array of objects into a new array of relations conditioned
on a contextual feature. The unit computes sparse high-order relations
between the input objects, and then modulates the encoding through
a specified context (See Fig.~\ref{fig:Illustration-of-Multiscales}).
The flexibility of $\UnitName$ and its encapsulating design allow
it to be replicated and layered to form deep hierarchical conditional
relation networks ($\ModelName$) in a straightforward manner (See
Fig.\ \ref{fig:visual_stream} and Fig.\ \ref{fig:textual_stream}).
Within the scope of this work, the $\ModelName$ is a two-stream network
of visual content and textual subtitles, in which the two sub-networks
share a similar design philosophy but are customized to suit the properties
of each input modality. Whilst the visual stream is built up by stacked
$\UnitName$ units at different granularities, the textual stream
is composed of a single $\UnitName$ unit taking linguistic segments
as inputs. The stacked units in the visual stream thus provide contextualized
refinement of relational knowledge from visual objects -- in a stage-wise
manner it combines appearance features with clip activity flow and
linguistic context, and afterwards incorporates the context information
from the whole video motion and linguistic features. On the textual
side, the $\UnitName$ unit functions in the same manner but on textual
objects. The resulting $\ModelName$ is homogeneous, agreeing with
the design philosophy of networks such as InceptionNet \cite{szegedy2015going},
ResNet \cite{he2016deep} and FiLM \cite{perez2018film}.

The hierarchy of the $\UnitName$s for each input modality is shown
as follows. At the lowest level of the visual stream, the $\UnitName$s
encode the relations \emph{between} frame appearance in a clip and
integrate the \emph{clip motion as context}; this output is processed
at the next stage by $\UnitName$s that now integrate in the \emph{linguistic
context}. In the following stage, the $\UnitName$s capture the relation
\emph{between} the clip encodings, and integrate in \emph{video motion
as context}. In the final stage the $\UnitName$ integrates the video
encoding with the linguistic feature as context (See Fig.~\ref{fig:visual_stream}).
As for the textual stream, due to its high-level abstraction as well
as the diversity of nuances of expression comparing to its visual
counterpart, we only use the $\UnitName$ to encode relations between\emph{
}linguistic segments in a given dialogue extracted from textual subtitles
and leverage well-known techniques in sequential modeling, such as
LSTM \cite{hochreiter1997long} or BERT \cite{devlin2018bert}, to
understand sequential relations at the word level. By allowing the
$\UnitName$s to be stacked hierarchically, the model naturally supports
modeling hierarchical structures in video and relational reasoning.
Likewise, by allowing appropriate context to be introduced in stages,
the model handles multimodal fusion and multi-step reasoning. For
long videos, further levels of hierarchy can be added enabling encoding
of relations between distant frames.

We demonstrate the capability of $\ModelName$ in answering questions
in major $\Problem$ datasets, including both short-form and long-form
videos. The hierarchical architecture with four-layers of $\UnitName$
units achieves favorable answer accuracy across all $\Problem$ tasks.
Notably, it performs consistently well on questions involving either
appearance, motion, state transition, temporal relations, or action
repetition demonstrating that the model can analyze and combine information
in all of these channels. Furthermore the $\ModelName$ scales well
on longer length videos simply with the addition of an extra layer.
Fig.~\ref{fig:Examples-of-short-form} and Fig.~\ref{fig:Examples-of-long-form}
demonstrate several representative cases those were difficult for
the baseline of flat visual-question interaction but can be handled
by our model. Our model and results demonstrate the impact of building
general-purpose neural reasoning units that support native multimodality
interaction in improving robustness and generalization capacities
of $\Problem$ models.

This paper advances the preliminary work \cite{le2020hierarchical}
in three main aspects: (1) devising a new network architecture that
leverages the design philosophy of HCRN in handling long-form Video
QA, demonstrating the flexibility and generic applicability of the
proposed CRN unit in various input modalities; (2) providing a comprehensive
theoretical analysis on the complexity of CRN computational unit as
well as the resulting HCRN architectures; (3) conducting more rigorous
experiments and ablation study to fully examining the capability of
the proposed network architectures, especially for the long-form Video
QA. These experiments thoroughly assure the consistency in behavior
of the CRN unit upon different settings and input modalities. We also
simplify the sampling procedure in the main algorithm, reducing the
run time significantly.

The rest of the paper is organized as follows. Section~\ref{sec:Related-Work}
reviews related work. Section~\ref{sec:Method} details our main
contributions -- the CRN, the HCRNs for Video QA on both short-from
videos and long-form videos that include subtitles, as well as complexity
analysis. The next section describes the results of the experimental
suite. Section~\ref{sec:Discussion} provides further discussion
and concludes the paper.

\section{Related work \label{sec:Related-Work}}

Our proposed $\ModelName$ model advances the development of $\Problem$
by addressing three key challenges: (1) Efficiently representing videos
as amalgam of complementing factors including appearance, motion and
relations, (2) Effectively allows the interaction of such visual features
with the linguistic query and (3) Allows integration of different
input modalities in $\Problem$ with one unique building block.

\textbf{Long/short-form Video QA} is a natural extension of image
QA and it has been gathering a rising attention in recent years, with
the release of a number of large-scale short-form Video QA datasets,
such as TGIF-QA \cite{jang2017tgif,xu2016msr}, as well as long-form
MovieQA datasets with accompanying textual modality, such as TVQA
\cite{lei2018tvqa}, MovieQA \cite{tapaswi2016movieqa} and PororoQA
\cite{kim2017deepstory}. All studies on $\Problem$ treats short-form
and long-form $\Problem$ as two separate problems in which proposed
methods \cite{fan2019heterogeneous,gao2018motion,jang2017tgif,kim2017deepstory,kim2019progressive,lei2018tvqa}
are deviated to handle either one of the two problems. Different from
those works, this paper takes the challenge of designing a generic
method that covers both long-form and short-form $\Problem$ with
simple exercises of block stacking and rearrangements to switch one
unique model between the two problem depending upon the availability
of input modalities for each of them.

\textbf{Spatio-temporal video representation} is traditionally done
by variations of recurrent networks (RNNs) among which many were used
for $\Problem$ such as recurrent encoder-decoder \cite{zhu2017uncovering,zhao2019long},
bidirectional LSTM \cite{kim2017deepstory} and two-staged LSTM \cite{zeng2017leveraging}.
To increase the memorizing ability, external memory can be added to
these networks \cite{gao2018motion,zeng2017leveraging}. This technique
is more useful for videos that are longer \cite{xu2016msr} and with
more complex structures such as movies \cite{tapaswi2016movieqa}
and TV programs \cite{lei2018tvqa} with extra accompanying channels
such as speech or subtitles. On these cases, memory networks \cite{kim2017deepstory,na2017read,wang2019holistic}
were used to store multimodal features \cite{wang2018movie} for later
retrieval. Memory augmented RNNs can also compress video into heterogeneous
sets \cite{fan2019heterogeneous} of dual appearance/motion features.
While in RNNs, appearance and motion are modeled separately, 3D and
2D/3D hybrid convolutional operators \cite{tran2018closer,qiu2017learning}
intrinsically integrates spatio-temporal visual information and are
also used for $\Problem$ \cite{jang2017tgif,li2019beyond}. Multiscale
temporal structure can be modeled by either mixing short and long
term convolutional filters \cite{wu2019long} or combining pre-extracted
frame features non-local operators \cite{tang2018non,li2017temporal}.
Within the second approach, the TRN network \cite{zhou2018temporal}
demonstrates the role of temporal frame relations as an another important
visual feature for video reasoning and Video QA \cite{le2neural}.
Relations of predetected objects were also considered in a separate
processing stream \cite{jin2019multi} and combined with other modalities
in late-fusion \cite{singh2019spatio}. Our $\ModelName$ model emerges
on top of these trends by allowing all three channels of video information
namely appearance, motion and relations to iteratively interact and
complement each other in every step of a hierarchical multi-scale
framework.

Earlier attempts for generic multimodal fusion for visual reasoning
include bilinear operators, either applied directly \cite{kim2018bilinear}
or through attention \cite{kim2018bilinear,yu2017multi}. While these
approaches treat the input tensors equally in a costly joint multiplicative
operation, $\ModelName$ separates conditioning factors from refined
information, hence it is more efficient and also more flexible on
adapting operators to conditioning types.

Temporal hierarchy has been explored for video analysis \cite{lienhart1999abstracting},
most recently with recurrent networks \cite{pan2016hierarchical,baraldi2017hierarchical}
and graph networks \cite{mao2018hierarchical}. However, we believe
we are the first to consider hierarchical interaction of multi-modalities
including linguistic cues for $\Problem$.

\textbf{Linguistic query--visual feature interaction} \textbf{in
Video QA} has traditionally been formed as a visual information retrieval
task in a common representation space of independently transformed
question and referred video \cite{zeng2017leveraging}. The retrieval
is more convenient with heterogeneous memory slots \cite{fan2019heterogeneous}.
On top of information retrieval, co-attention between the two modalities
provides a more interactive combination \cite{jang2017tgif}. Developments
along this direction include attribute-based attention \cite{ye2017video},
hierarchical attention \cite{liang2018focal,zhao2018multi,zhao2017video},
multi-head attention \cite{kim2018multimodal,li2019learnable}, multi-step
progressive attention memory \cite{kim2019progressive} or combining
self-attention with co-attention \cite{li2019beyond}. For higher
order reasoning, question can interact iteratively with video features
via episodic memory or through switching mechanism \cite{yang2019question}.
Multi-step reasoning for Video QA is also approached by \cite{xu2017video}
and \cite{song2018explore} with refined attention.

Unlike these techniques, our $\ModelName$ model supports conditioning
video features with linguistic clues as a context factor in every
stage of the multi-level refinement process. This allows the representation
of linguistic cues to be involved earlier and deeper into video presentation
construction than any available methods.

\textbf{Neural building blocks} - Beyond the Video QA domain, $\UnitName$
unit shares the idealism of uniformity in neural architecture with
other general purpose neural building blocks such as the block in
InceptionNet \cite{szegedy2015going}, Residual Block in ResNet \cite{he2016deep},
Recurrent Block in RNN, conditional linear layer in FiLM \cite{perez2018film},
and matrix-matrix-block in neural matrix net \cite{do2018learning}.
Our $\UnitName$ departs significantly from these designs by assuming
an array-to-array block that supports conditional relational reasoning
and can be reused to build networks of other purposes in vision and
language processing. As a result, our $\ModelName$ is a perfect fit
not only for short-form video where questions are all about visual
content of a video snippet but also for long-form video (Movie QA)
where a model has to look at both visual cue and textual cue (subtitles)
to arrive at correct answers. Due to the great challenges posed by
long-form $\Problem$ and the diversity in terms of model building,
current approaches in $\Problem$ mainly spend efforts on handling
the visual part while leaving the textual part for common techniques
such as LSTM \cite{lei2018tvqa} or latest advance in natural language
processing BERT \cite{yang2020bert}. To the best of our knowledge,
$\ModelName$ is the first model that could solve both short-form
and long-form $\Problem$ with models building up from a generic neural
block.

\section{Method \label{sec:Method}}

\begin{figure}
\begin{centering}
\includegraphics[width=0.95\columnwidth]{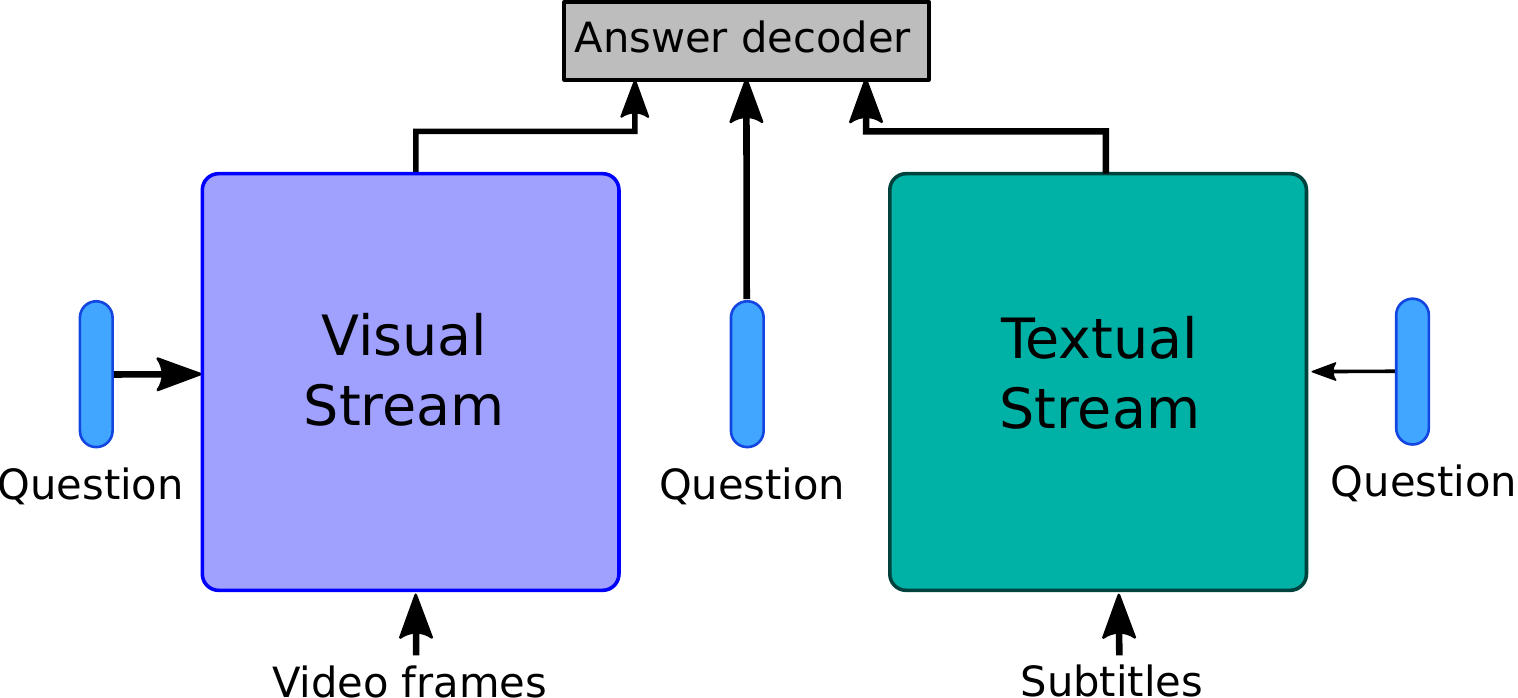}
\par\end{centering}
\caption{Overall multimodal Video QA architecture. The two streams handle visual
and textual modalities in parallel, followed by an answer decoder
for feature joining and prediction.\label{fig:Overall-VideoQA}}
\end{figure}

The goal of $\Problem$ is to deduce an answer $\tilde{a}$ from a
video $\mathcal{V}$ and optionally from additional information channels
such as subtitles $\mathcal{S}$ in response to a natural question
$q$. The answer $\tilde{a}$ can be found in an answer space $\mathcal{A}$
which is a pre-defined set of possible answers for open-ended questions
or a list of answer candidates in the case of multi-choice questions.
Formally, $\Problem$ can be formulated as follows:

\begin{equation}
\tilde{a}=\underset{a\in\mathcal{A}}{\text{argmax}}\mathcal{F}_{\theta}\left(a\mid q,\mathcal{V},\mathcal{S}\right),\label{eq:prob-def}
\end{equation}
where $\theta$ is the model parameters of scoring function $\mathcal{F}$.

Within the scope of this work, we address two common settings of $\Problem$:
(a) short-form $\Problem$ where the visual content of a given single
video shot singularly suffice to answer the questions and (b) long-form
$\Problem$ where the essential information disperses among visual
content in the multi-shot video sequence and conversational content
in accompanying textual subtitles.

$\ModelName$ was designed in the endeavor for a homogeneous neural
architecture that can adapt to solve both problems. Its overall workflow
is depicted in Fig.~\ref{fig:Overall-VideoQA}. In long-form videos,
when both visual and textual streams are present, $\ModelName$ distills
relevant information from the visual stream and the textual stream,
both are conditioned on the question. Eventually, it combines them
into an answer decoder for final prediction in late-fusion multimodal
integration. In short-form cases, where only video frames are available,
the visual stream is solely active, working with the single-input
answer decoder. One of the ambitions of the design is to build each
processing stream as a hierarchical network simply by stacking common
core processing units of the same family. Similar to all previous
deep-learning-based approaches in the literature \cite{jang2017tgif,li2019beyond,fan2019heterogeneous,gao2018motion,lei2018tvqa,le2neural},
our HCRN operates on top of the feature embeddings of multiple input
modalities, making use of the powerful feature representations extracted
by either common visual recognition models pre-trained on large-scale
datasets such as ResNet \cite{he2016deep}, ResNeXt \cite{xie2017aggregated,hara2018can}
or pre-trained word embeddings such as GloVe \cite{pennington2014glove}
and BERT \cite{devlin2018bert}.

In the following subsections, we present the design of the core unit
in Sec.~\ref{subsec:Relation-Network}, the hierarchical designs
tailored to each modality in Sec.~\ref{subsec:HCRN}, the answer
decoder in Sec.~\ref{subsec:Answer-Decoders}. Theoretical analysis
of the computational complexity of the models follows in Sec.~\ref{subsec:Complexity-Analysis}.

\subsection{Conditional relation network unit\label{subsec:Relation-Network}}

\begin{algorithm}[t]
\small
\label{algo:CRN}
\caption{CRN Unit}
	\SetKwInOut{Input}{Input}
	\SetKwInOut{Output}{Output}
	\SetKwInOut{Metaparams}{Metaparams}
	\Input{Array $\mathcal{X}=\{x_i\}_{i=1}^n$, conditioning feature $c$}
	\Output{Array $R$}
	\Metaparams{$\{k_{\textrm{max}},t \mid k_{\textrm{max}}<n\}$}
	
	Initialize $R \leftarrow \{\}$

	\For{$k \leftarrow 2$ \KwTo $k_{\SI{}{max}}$}
	{
		$Q^{k}=$ randomly select $t$ subsets of size $k$ from $\mathcal{X}$
		
		\For{\SI{}{\textbf{each}} subset $q_i$ $\in$ $Q^{k}$}
		{	
			$g_i = g^k(q_i)$
			
			$h_i = h^k(g_i,c)$
		}
        
		$r^k=p^{k}(\{h_i\})$
		
		add $r^k$ to $R$
		
	}
\end{algorithm}

\begin{figure}
\begin{centering}
\includegraphics[width=0.95\columnwidth]{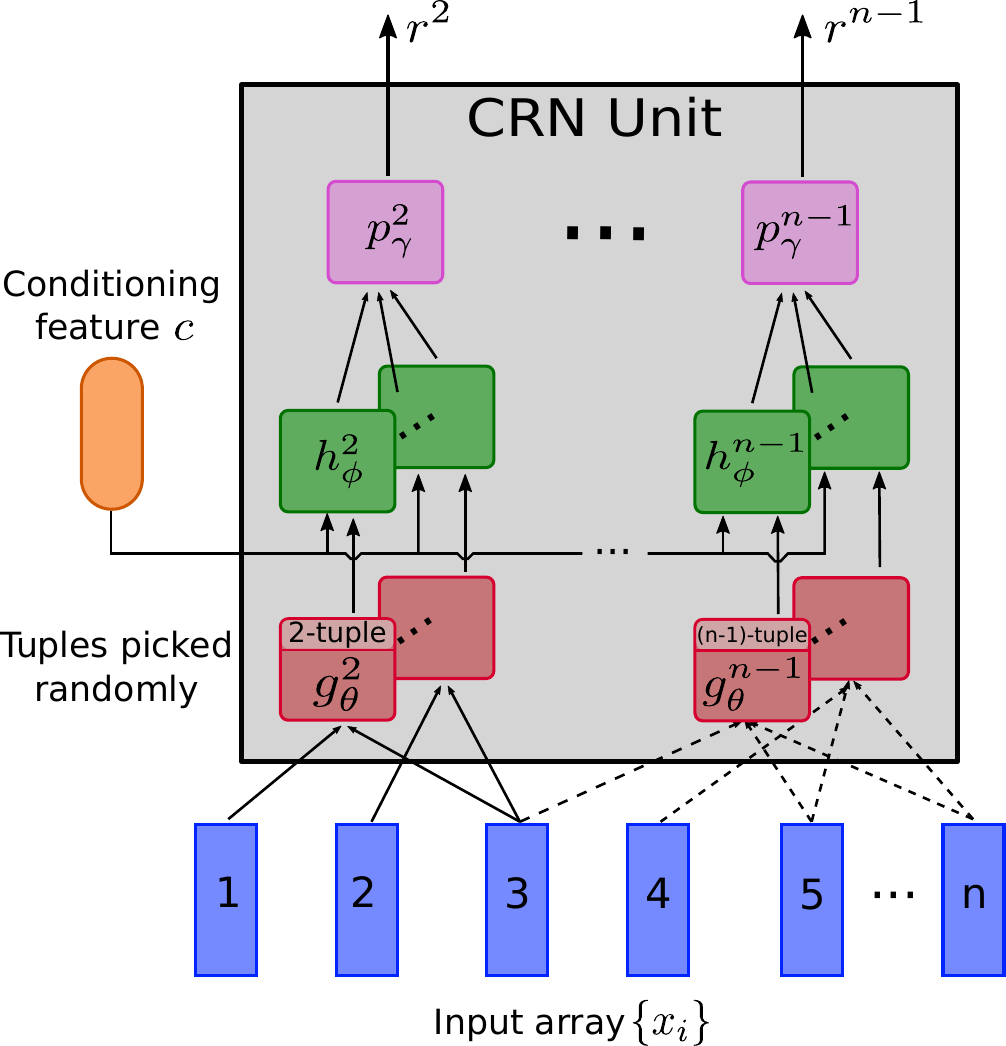}
\par\end{centering}
\caption{Conditional Relation Network. a) Input array {\small{}$\mathcal{X}$}
of $n$ objects are first processed to model $k$-tuple relations
from $t$ sub-sampled size-$k$ subsets by sub-network $g^{k}(.)$.
The outputs are further conditioned with the context $c$ via sub-network
$h^{k}(.,.)$ and finally aggregated by $p^{k}(.)$ to obtain a result
vector $r^{k}$ which represents $k$-tuple conditional relations.
Tuple sizes can range from $2$ to $(n-1)$, which outputs an $(n-2)$-dimensional
output array.\label{fig:Illustration-of-Multiscales}}
\medskip{}
\end{figure}

\begin{table}
\begin{centering}
\begin{tabular}{|l|l|}
\hline 
Notation & Role\tabularnewline
\hline 
\hline 
$\mathcal{X}$ & Input array of $n$ objects (e.g. frames, clips)\tabularnewline
\hline 
$c$ & Conditioning feature (e.g. query, motion feat.)\tabularnewline
\hline 
$k_{\text{max}}$ & Maximum subset (also tuple) size considered\tabularnewline
\hline 
$k$ & Each subset size from $2$ to $k_{max}$\tabularnewline
\hline 
$t$ & Number of size-$k$ subsets of $\mathcal{X}$ randomly sampled\tabularnewline
\hline 
$Q^{k}$ & Set of $t$ size-$k$ subsets sampled from $\mathcal{X}$\tabularnewline
\hline 
$g^{k}(.)$ & Sub-network processing each size-$k$ subset\tabularnewline
\hline 
$h^{k}(.,.)$ & Conditioning sub-network\tabularnewline
\hline 
$p^{k}(.)$ & Aggregating sub-network\tabularnewline
\hline 
$R$ & Result array of CRN unit on $\mathcal{X}$ given $c$\tabularnewline
\hline 
$r^{k}$ & Member result vector of $k$-tuple relations\tabularnewline
\hline 
\end{tabular}\medskip{}
\par\end{centering}
\centering{}\caption{Notations of CRN unit operations.\label{tab:notations}}
\end{table}

We introduce a general computation unit, termed Conditional Relation
Network ($\UnitName$), which takes as input an array of $n$ objects
$\mathcal{X}=\left\{ x_{i}\right\} _{i=1}^{n}$ and a conditioning
feature $c$ serving as global context. Objects are assumed to live
either in the same vector space $\mathbb{R}^{d}$ or tensor space,
for example, $\mathbb{R}^{W\times H\times d}$ in case of images (or
video frames). CRN generates an output array of objects of the same
dimensions containing high-order object relations of input features
given the global context. The global context is problem-specific,
serving as a modulator to the formation of the relations. When in
use for $\Problem$, CRN's input array is composed of features at
either frame level, short-clip levels, or textual features. Examples
of global context include the question and motion profiles at a given
hierarchical level. 

 Given input set of object $\mathcal{X},$ CRN first considers the
set of subsets $\mathcal{Q}=\left\{ Q^{k}\right\} _{k=2}^{k_{\textrm{max}}}$
where each set $Q^{k}$ contains $t$ size-$k$ subsets randomly sampled
from $\mathcal{X}$, where $t$ is the sampling frequency, $t<C(n,k)$.
On each collection $Q^{k}$, $\UnitName$ then uses member relational
sub-networks to infer the joint representation of $k$-tuple object
relations. In videos, due to temporal coherence, the objects $\left\{ x_{i}\right\} _{i=1}^{n}$
share a great amount of mutual information, therefore, it is reasonable
to use a subsampled set of $t$ combinations instead of considering
all possible combinations for $Q^{k}$. This is inspired by \cite{zhou2018temporal},
however, we sample $t$ size-$k$ subsets directly from the original
$\mathcal{\mathcal{X}}$ rather than randomly take subsets in a pool
of all possible size-$k$ subsets as in their method. By doing this,
we can reduce the computational complexity of the $\UnitName$ as
it is much cheaper to sample elements out of a set than building all
combinations of subsets which cost $\mathcal{O}(2^{n})$ time. We
provide more analysis on the complexity of the $\UnitName$ unit later
in Sec. \ref{subsec:Complexity-Analysis}.

Regarding relational modeling, each member subset of $Q^{k}$ then
goes through a set-based function $g^{k}(.)$ for relational modeling.
The relations across objects are then further refined with a conditioning
function $h^{k}(.,c)$ in the light of conditioning feature $c$.
Finally, the $k$-tuple relations are summarized by the aggregating
function $p^{k}(.)$. The similar operations are done across the range
of subset size from 2 to $k_{\text{max}}$. Regarding the choice of
$k_{\text{max}}$, we use $k_{\text{max}}=n-1$ in later experiments,
resulting in the output array of size $n-2$ if $n>2$ and an array
of size $1$ if $n=2$.

The detailed operation of the CRN unit is presented formally as pseudo-code
in Alg.~1 and visually in Fig.~\ref{fig:Illustration-of-Multiscales}.
Table~\ref{tab:notations} summarizes the notations used across these
presentations.

\subsubsection{Networks implementation}

\paragraph{Set aggregation:}

The set functions $g^{k}(.)$ and $p^{k}(.)$ can be implemented as
any aggregation sub-networks that join a random set into a single
representation. As a choice in implementation, the function $g^{k}(.)$
is either average pooling or a simple concatenation operator while
$p^{k}(.)$ is average pooling.

\paragraph{Conditioning function:}

The design of the conditioning sub-network that implements $h^{k}(.,c)$
depends on the relationship between the input set $\mathcal{X}$ and
the conditioning feature $c$ as well as the properties of the channels
themselves. Here we present four forms of neural operation implementing
this function.

\textbf{Additive form:}

A simple form of $h^{k}(.,c)$ is feature concatenation followed by
a MLP that models the non-linear relationships between multiple input
modalities:

\begin{equation}
h^{k}\left(.,c\right)=\text{ELU}\left(W_{h_{1}}\left[.;c\right]\right),\label{eq:conditioning_concat}
\end{equation}
where $[\thinspace;]$ to denote the tensor concatenation operation
and ELU is the non-linear activation function introduced in \cite{clevert2015fast}.
Eq.~\ref{eq:conditioning_concat} is sufficient when $x$ and $c$
are additively complementary. 

\textbf{Multiplicative form:}

To support more complex the relationship between the input $x$ and
the conditioning feature $c$, a more sophisticated joining operation
is warranted. For example, when $c$ implies a selection criterion
to modulate the relationship between elements in $x$, the multiplicative
relation between them can be represented in conditioning function
by:

\begin{equation}
h^{k}\left(x,c\right)=\text{ELU}\left(W_{h_{1}}\left[x;x\odot c;c\right]\right),\label{eq:conditioning_mul}
\end{equation}
where $\odot$ denotes Hadamard product.

\textbf{Sequential form:}

As aforementioned, how to properly choose the sub-network $h^{k}(.,c)$
is also driven by the properties of the input set $\mathcal{X}$.
Given the context of $\Problem$ of later use of the $\UnitName$
unit where elements in $\mathcal{X}$ may contain strong temporal
relationships, we additionally integrate sequential modeling capability,
which is a BiLSTM network in this paper, along with the conditioning
sub-network as presented in Eq.~\ref{eq:conditioning_concat} and
Eq.~\ref{eq:conditioning_mul}. Formally, $h^{k}(.,c)$ is defined
as:

\begin{align}
s & =\left[x;x\odot c;c\right],\label{eq:conditioning}\\
s^{\prime} & =\text{BiLSTM}(s),\label{eq:conditioning_bilstm}\\
h^{k}(.,c) & =\text{maxpool}(s^{\prime}).\label{eq:max_temporally}
\end{align}

\textbf{Dual conditioning form:}

In the later use of $\UnitName$ in $\Problem$ where it can happen
that two concurrent signals $c_{1},c_{2}$ are used as conditioning
features, we simply extend Eq.~\ref{eq:conditioning_concat} and
Eq.~\ref{eq:conditioning_mul} and Eq.~\ref{eq:conditioning} as:
\begin{align}
h^{k}\left(x,c\right)= & \text{\ ELU}\left(W_{h_{1}}\left[x;c_{1};c_{2}\right]\right),\\
h^{k}\left(x,c\right)= & \text{\ ELU}\left(W_{h_{1}}\left[x;x\odot c_{1};x\odot c_{2};c_{1};c_{2}\right]\right),\\
s= & \left[x;x\odot c_{1};x\odot c_{2};c_{1};c_{2}\right],
\end{align}
respectively.

\subsection{Hierarchical conditional relation network for multimodal Video QA
\label{subsec:HCRN}}

We use $\UnitName$ blocks to build a deep network architecture that
supports a wide variety of $\Problem$ settings. In particular, two
variations are specifically designed to work on short-form and long-form
$\Problem$. For each of the settings, the network design adapts to
exploit inherent characteristics of a video sequence namely temporal
relations, motion, linguistic conversation and the hierarchy of video
structure, and to support reasoning guided by linguistic questions.
We term the proposed network architecture Hierarchical Conditional
Relation Networks ($\ModelName$). The design of the $\ModelName$
by stacking reusable core units is partly inspired by modern CNN network
architectures, of which InceptionNet \cite{szegedy2015going} and
ResNet \cite{he2016deep} are the most well-known examples. In the
general form of $\Problem$, $\ModelName$ is a multi-stream end-to-end
differentiable neural network in which one stream is to handle visual
content and the other one handling textual dialogues in subtitles.
The network is modular and each network stream plays a plug-and-play
role adaptively to the presence of input modalities.

\subsubsection{Preprocessing\label{subsec:Preprocessing}}

With the $\Problem$ as defined in Eq. \ref{eq:prob-def}, HCRN takes
input and question represented as sets of visual or textual objects
and computes the answer. In this section, we describe the preprocessing
of raw videos into appropriate input sets for $\ModelName$.

\paragraph{Visual representation:}

We begin by dividing the video $\mathcal{V}$ of $L$ frames into
$N$ equal length clips $C=\{C_{1},...,C_{N}\}$. For short-form videos,
each clip $C_{i}$ of length $T=\left\lfloor L/N\right\rfloor $ is
represented by two sources of information: frame-wise appearance feature
vectors $V_{i}=\left\{ v_{i,j}\mid v_{i,j}\in\mathbb{R^{\text{2048}}}\right\} _{j=1}^{T}$
, and a motion feature vector at clip level $f_{i}\in\mathbb{R^{\text{2048}}}$.
Appearance features are vital for video understanding as the visual
saliency of objects/entities in the video is usually of interest to
human questions. In short clips, moving objects and events are primary
video artifacts that capture our attention. Hence, it is common to
see the motion features coupled with the appearance features to represent
videos in the video understanding literature. On the contrary, in
long-form video such as those in movies and TV programs, the concerns
can be less about specific motions but more into movie plot or film
grammar. As a result, we use the frame-wise appearance as the only
feature for the long-form $\Problem$. In our experiments, $v_{i,j}$
are the \emph{pool5} output of ResNet \cite{he2016deep} features
and $f_{i}$ are derived by ResNeXt-101 \cite{xie2017aggregated,hara2018can}.

Subsequently, linear feature transformations are applied to project
$\{v_{ij}\}$ and $f_{i}$ into a standard $d$-dimensions feature
space to obtain $\left\{ \hat{v}_{ij}\mid\hat{v}_{ij}\in\mathbb{R^{\text{d}}}\right\} $
and $\hat{f}_{i}\in\mathbb{R^{\text{d}}}$, respectively.

\paragraph{Linguistic representation:}

Linguistic objects are built from question, answer choices and long-form
videos' subtitles. We explore two options of representation learning
for them using BiLSTM and BERT.
\begin{itemize}
\item Sequential embedding with BiLSTM:
\end{itemize}
All words in linguistic cues including those in questions, answer
choices and subtitles are first embedded into vectors of 300 dimensions
by pre-trained GloVe word embeddings \cite{pennington2014glove}. 

For question and answer choices, we further pass these context-independent
embedding vectors through a BiLSTM. Output hidden states of the forward
and backward LSTM passes are finally concatenated to form the overall
query representation $q\in\mathbb{R}^{d}$ for the questions and $a\in\mathbb{R}^{d}$
for the answer choices if available (multi-choice question-answer
pairs). 

For the accompanying subtitles provided in long-form $\Problem$,
instead of treating them as one big passage as in prior works \cite{lei2018tvqa,kim2019progressive},
we dissect the subtitle passage $\mathcal{S}$ into a fixed number
of $M$ overlapping \emph{segments} $\mathcal{U}=\{\mathcal{U}_{1},..,\mathcal{U}_{M}\}$.
The number of words in sibling segments is identical $T=\textrm{length}(\mathcal{S})/M$
but varies from one video to another depending on the overall length
of the given subtitle passage $\mathcal{S}$. 

Similarly to the case of processing questions, we process each segment
with pre-trained word embeddings followed by a BiLSTM. The hidden
states of the BiLSTM which is also of size $T$ are then used as textual
objects:
\begin{equation}
U_{i}=\textrm{BiLSTM}(\mathcal{U}_{i}).
\end{equation}
$\left\{ U_{i}\right\} _{i=1}^{M}$ are the final representations
ready to be used by the textual stream which will be described in
Sec.~\ref{subsec:Textual-stream}.
\begin{itemize}
\item Contextual embedding using BERT:
\end{itemize}
As an alternative option for linguistic representation, we utilize
pre-trained BERT network \cite{devlin2018bert} to extract contextual
word embeddings. Instead of encoding words independently as in GloVe,
BERT embeds each word in the context of its surrounding words using
a self attention mechanism.

For short-form $\Problem$, we tokenize the given question and answer
choices in multi-choice questions and subsequently feed the tokens
into the pre-trained BERT network. Averaged embeddings of words in
a sentence are used as a unified representation for that sentence.
This applies to generate both the question representation $q$ and
answer choices $\{a_{i}\}_{i=1,...,A}$.

For long-form $\Problem$, with each answer choice, we form a long
string $l_{i}$ by stacking it with subtitles $\mathcal{S}$ and the
question sentence. We then tokenize and embed each string $l_{i}$
with BERT into contextual hidden matrix $H$ of size $m\times d$
where $m$ is the maximum number of tokens in the input and $d$ is
hidden dimensions of BERT. This tensor $H$ is then split up into
corresponding embedding vectors of the subtitles $H^{s}$, of the
question $H^{q}$ and of the answer choice $H^{a_{i}}$:
\begin{equation}
\left(H^{s},H^{q},H^{a_{i}}\right)=\textrm{BERT}(l_{i}).
\end{equation}

Eventually, we suppress the contextual tokens question representation
and answer choices into their respective single representation by
mean pooling:
\begin{equation}
q=\textrm{mean}(H^{q});\:a_{i}=\textrm{mean}(H^{a_{i}}),
\end{equation}
while keep those of subtitles $H^{s}$ as a set of textual objects.
$\left\{ U_{i}\right\} _{i=1}^{M}$ are obtained by sliding overlapping
windows of the same length over $H^{s}$.

\subsubsection{Visual stream\label{subsec:Visual-stream}}

\begin{figure*}
\begin{centering}
\includegraphics[width=0.65\textwidth]{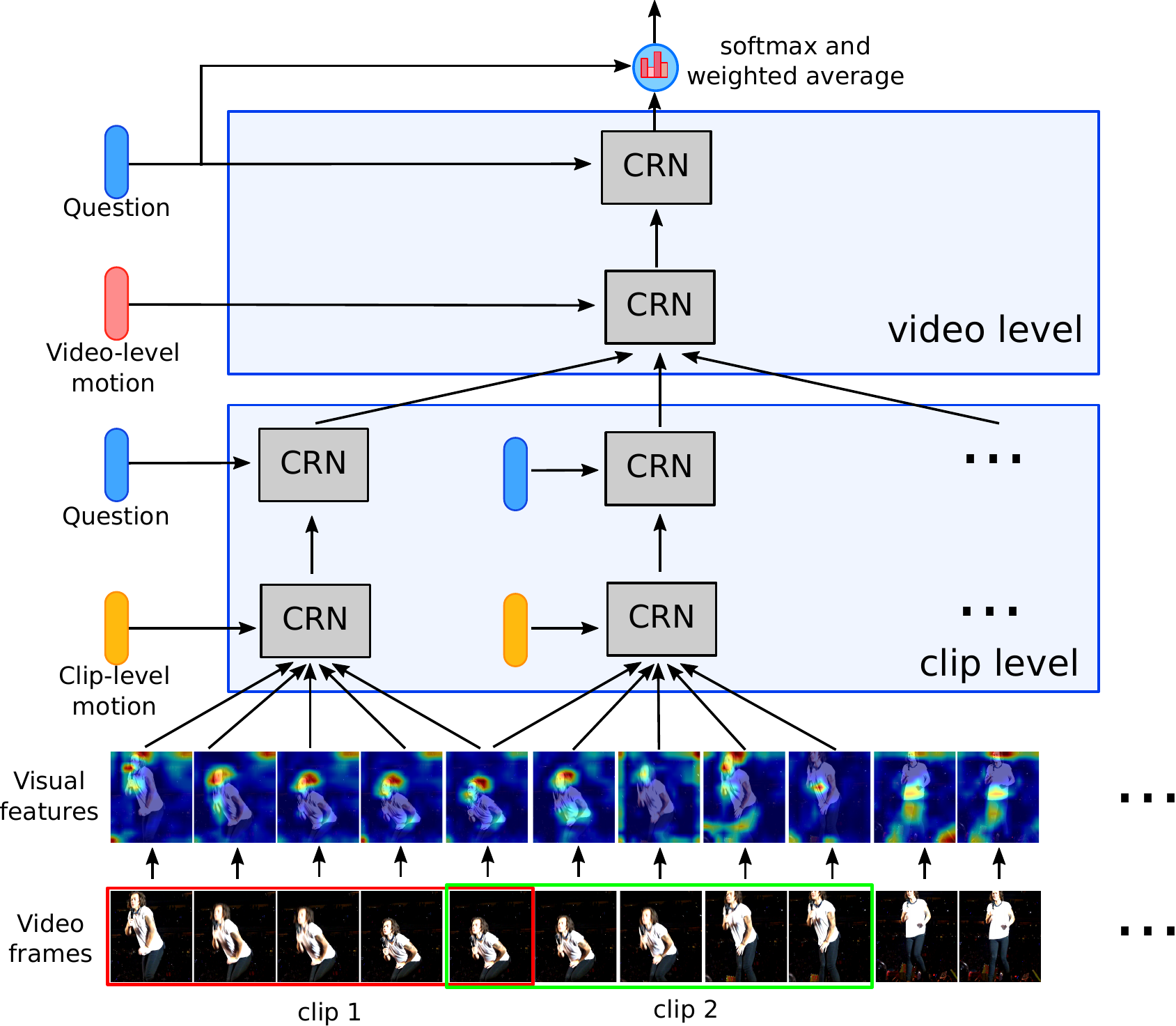}
\par\end{centering}
\caption{Visual stream. The $\protect\UnitName$s are stacked in a hierarchy,
embedding the video input at different granularities including frame,
short clip and entire video levels. At each level of granularity,
the video feature embedding is conditioned on the respective level-wise
motion feature and universal linguistic cue. \label{fig:visual_stream}}
\end{figure*}

An effective model for $\Problem$ needs to distill the visual content
in the context of the question and filter out the usually large portion
of the data that is not relevant to the question. Drawing inspiration
from the hierarchy of video structure, we boil down the problem of
$\Problem$ into a process of video representation in which a given
video is encoded progressively at different granularities, including
short clip (a sequence of video frames) and entire video levels (a
sequence of clips). It is crucial that \emph{the whole process conditions
on linguistic cues}.

With that in mind, we design a two-level structure to represent a
video, one at clip level and the other one at video level, as illustrated
in Fig\@.~\ref{fig:visual_stream}. At each hierarchy level, we
use two stacked $\UnitName$ units, one conditioned on motion features
followed by one conditioned on linguistic cues. Intuitively, the motion
feature serves as a dynamic context shaping the temporal relations
found among frames (at the clip level) or clips (at the video level).
It also provides a saliency indicator of which relations are worth
the attention \cite{mahapatra2008motion}.

As the shaping effect is applied to all relations in a complementary
way, selective (multiplicative) relation between the relations and
the conditioning feature is not needed, and thus a simple concatenation
operator followed by a MLP suffices. On the other hand, the linguistic
cues are by nature selective, that is, not all relations are equally
relevant to the question. Thus we utilize the multiplicative form
for feature fusion as in Eq.~\ref{eq:conditioning_mul} for the $\UnitName$
units which condition on question representation. 

With this particular design of network architecture, the input array
at clip level consists of frame-wise appearance feature vectors $\{\hat{v}_{ij}\}$,
while that at a video level is the output at the clip level. The
motion conditioning feature at clip level CRNs is corresponding clip
motion feature vector $\hat{f}_{i}$. They are further passed to an
LSTM, whose final state is used as video-level motion features. Note
that this particular implementation is not the only option. We believe
we are the first to progressively incorporate multiple modalities
of input in such a hierarchical manner in contrast to the typical
approach of treating appearance features and motion features as a
two-stream network.

\paragraph{Deeper hierarchy:}

To handle a video of longer size, up to thousands of frames which
is equivalent to dozens of short-term clips, there are two options
to reduce the computational cost of $\UnitName$ in handling large
sets of subsets $\left\{ Q^{k}\mid k=2,...,k_{\text{max}}\right\} $
given an input array $\mathcal{X}$: (i) limit the maximum subset
size $k_{\text{max}}$, or (ii) extend the visual stream networks
to deeper hierarchy. For the former option, this choice of sparse
sampling may have the potential to lose critical relation information
of specific subsets. The latter, on the other hand, is able to densely
sample subsets for relation modeling. Specifically, we can group $N$
short-term clips into $N_{1}\times N_{2}$ hyper-clips, of which $N_{1}$
is the number of the hyper-clips and $N_{2}$ is the number of short-term
clips in one hyper-clip. By doing this, the visual stream now becomes
a 3-level of hierarchical network architecture. See Sec.~\ref{subsec:Complexity-Analysis}
for the effect of going deeper on running time and Sec.~\ref{subsec:Deepening-model-hierarchy}
on accuracy.

\paragraph{Computing the output:}

At the end of the visual stream, we compute the average visual feature
which is driven by the question representation $q$. Assuming that
the outputs of the last $\UnitName$ unit at video level are an array
$O=\left\{ o_{i}\mid o_{i}\in\mathbb{R}^{H\times d}\right\} _{i=1}^{N-4}$,
we first stack them together, resulting in an output tensor $o\in\mathbb{R}^{(N-4)\times H\times d}$.
We further vectorize this output tensor to obtain the final output
$o^{\prime}\in\mathbb{R}^{H^{\prime}\times d},H^{\prime}=(N-4)\times H$.
The weighted average information is given by:

\begin{align}
I & =\left[W_{o^{\prime}}o^{\prime};W_{o^{\prime}}o^{\prime}\odot W_{q}q\right],\label{eq:attention_vision}\\
I^{\prime} & =\text{ELU}\left(W_{I}I+b\right),\\
\gamma & =\text{softmax}\left(W_{I^{\prime}}I^{\prime}+b\right),\\
\tilde{o} & =\sum_{h=1}^{H^{\prime}}\gamma_{h}o_{h}^{\prime};\,\tilde{o}\in\mathbb{R}^{d},
\end{align}
where, $\left[.\ ,.\right]$ denotes concatenation operation, and
$\odot$ is the Hadamard product. In case of multi-choice questions
which are coupled with answer choices $\{a_{i}\}$, Eq.\ \ref{eq:attention_vision}
becomes $I=\left[W_{o^{\prime}}o^{\prime};W_{o^{\prime}}o^{\prime}\odot W_{q}q;W_{o^{\prime}}o^{\prime}\odot W_{a}a_{i}\right]$.

\subsubsection{Textual stream\label{subsec:Textual-stream}}

\begin{figure*}
\begin{centering}
\includegraphics[width=0.65\textwidth]{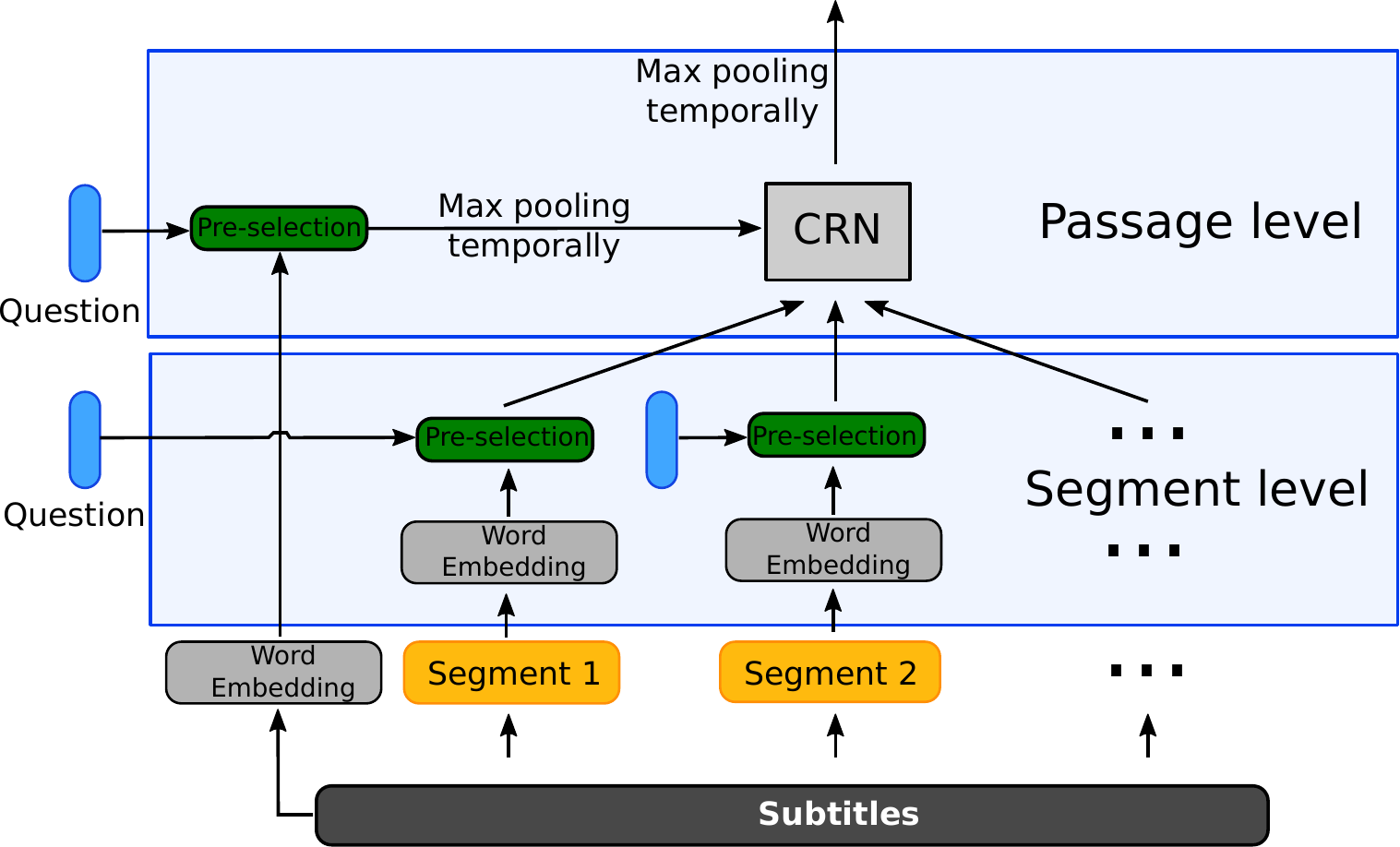}
\par\end{centering}
\caption{Textual stream. Both segment-level and passage-level textual objects
are modulated with the question by a \emph{pre-selection }module.
They then serves as input and conditioning features for an CRN modeling
long-term relationships between segments.\label{fig:textual_stream}}
\end{figure*}

$\ModelName$ architecture is readily applicable to the accompanying
textual subtitles in a similar bottom-up fashion as in visual stream.
The HCRN textual stream also consists of two-level hierarchy structure
that process textual objects of each segment and joining segments
into passage level (See Fig.~\ref{fig:textual_stream}).

The input of the stream is preprocessed subtitles, question, and answer
choices (Sec. \ref{subsec:Preprocessing}). The subtitles $\mathcal{S}$
is represented as its overall representation $H^{s}$ and also a sequence
of equal-length segments, each of which has been encoded into a set
of textual objects $U_{i}=\left\{ u_{i}^{t}\right\} _{t=1,..,T}\in\mathbb{R}^{T\times d}$.
Meanwhile, the question is encoded into a single vector $q\in\mathbb{R}^{d}$.

\paragraph{Question-relevant pre-selection:}

Unlike video frames, subtitles $\mathcal{S}$ contains irregularly
timed conversations between movie characters. Furthermore, while relevant
visual features are abundant throughout the video, only a small portion
of the subtitles is relevant to the query and reflective of the answers
$a$. To assure such relevance, antecedent to CRN unit, we modulate
the representation of passage $H^{s}$ and those of $M$ segments
$\left\{ U_{i}\right\} _{i=1}^{M}$ with both use the question and
the answer choice. It is done with the \emph{pre-selection} operator
described below.

At the segment level, the modulated representation $\hat{U}_{i}\in\mathbb{R}^{T\times d}$
of each segment $i$ of $T$ objects are produced by 
\begin{equation}
\hat{U}_{i}=W^{u}\left[U_{i};U_{i}\odot q\right].\label{eq:preselect_segment_level}
\end{equation}
Similarly, at video level, the subtitle modulated representation $\hat{H^{s}}\in\mathbb{R}^{S\times d}$
is built as

\begin{align}
\hat{H^{s}} & =W^{h}\left[H^{s};H^{s}\odot q\right].\label{eq:preselection_subtitle}
\end{align}

\paragraph{CRN unit:}

A single CRN unit of the stream operates at passage level which models
the relationships between segments. The modulated $\left\{ \hat{U}_{i}\right\} _{i=1}^{M}$
are passed as input objects to a $\UnitName$ unit. The conditioning
feature of the CRN is a max-pooled vector of the modulated representation
of the whole subtitle passage $\hat{H}^{s}$. At the end of the textual
stream, a temporal max-pooling operator is applied over the outputs
of the $\UnitName$ to obtain a single vector. 

\subsubsection{Adaptation \& implementation}

\paragraph{Short-form Video QA:}

Recall that short-form Video QA in this paper refers to QA about single-shot
videos of a few seconds without accompanying textual data. For these
cases, we employ the standard visual stream as described in Sec.~\ref{subsec:Visual-stream}
to distill video/question joint representation. This representation
is ready to be used by the answer decoder (See Sec.\ref{subsec:Answer-Decoders})
for generating the final output.

\paragraph{Long-form Video QA:}

Different from the short-form Video QA, long-form $\Problem$ involves
reasoning on both visual information from video frames and textual
content from subtitles. Compared to short snippets where local low-level
motion saliency plays a critical role, discovering high-level semantic
concepts associated with movie characters is more important \cite{sang2010character}.
Such semantics interleave in the data of both modalities. Further
difference comes from the fact that long-form videos are of greater
duration hence require appropriate treatment.

Although long-form videos share with short-forms in having a hierarchical
structure, they are distinctive in terms of semantic compositionality
and length. We employ visual and textual streams as described in Sec.~\ref{subsec:Visual-stream}
and Sec.~\ref{subsec:Textual-stream} with some adaptation for better
suitability with the data format and structure and simply use the
joint representation of the two streams for answer prediction. Ideally,
long-form $\Problem$ requires modeling interactions between a question,
visual content and textual content in subtitles. Whist both the visual
stream and textual stream described above involve early integration
of the question representation into a visual representation and textual
representation of subtitles, we do not opt for early interaction between
visual content and textual content in subtitles in this work. Pairwise
joint semantic representation between visual and language has proven
to be useful in $\Problem$ and closely related tasks \cite{yu2018joint},
however, it assumes the existence of pairs of multimodal sequence
data in which they are highly semantically compatible. Those pairs
are either a video sequence and a textual description of the video
or a video sequence and positive/negative answer choices to a question
about the visual content in the video. This is not always the case
for the visual content and textual content in subtitles in long-form
videos such as those taken from movies. Although the visual content
and textual content may complement each other to some extent, in many
cases, they may be greatly distant from each other. Let's take the
following scenario as an example, in a movie scene, two characters
are standing and chatting with each other. While the visual content
may provide information about where the conversation takes place,
it hardly contains any information about the topic of their conversation.
As a result, combining the visual information and textual information
at an early stage, in this case, has the potentials to cause information
distortion and make it difficult for information retrieval. In addition,
treating visual stream and textual stream in separation make it easier
to justify the benefits of using $\UnitName$ units in modeling the
relational information in each modality, hence, easier to assess the
generic use of the $\UnitName$ unit.

Note that in the visual stream in the $\ModelName$ architecture for
long-form $\Problem$, we use $\UnitName$ units to handle a subset
of sub-sampled frames in each clip at the clip level. At the video
level, another CRN gathers long-range dependencies between this clip-level
information sent up from lower level CRN outputs.

\begin{figure*}
\begin{centering}
\includegraphics[width=0.65\textwidth]{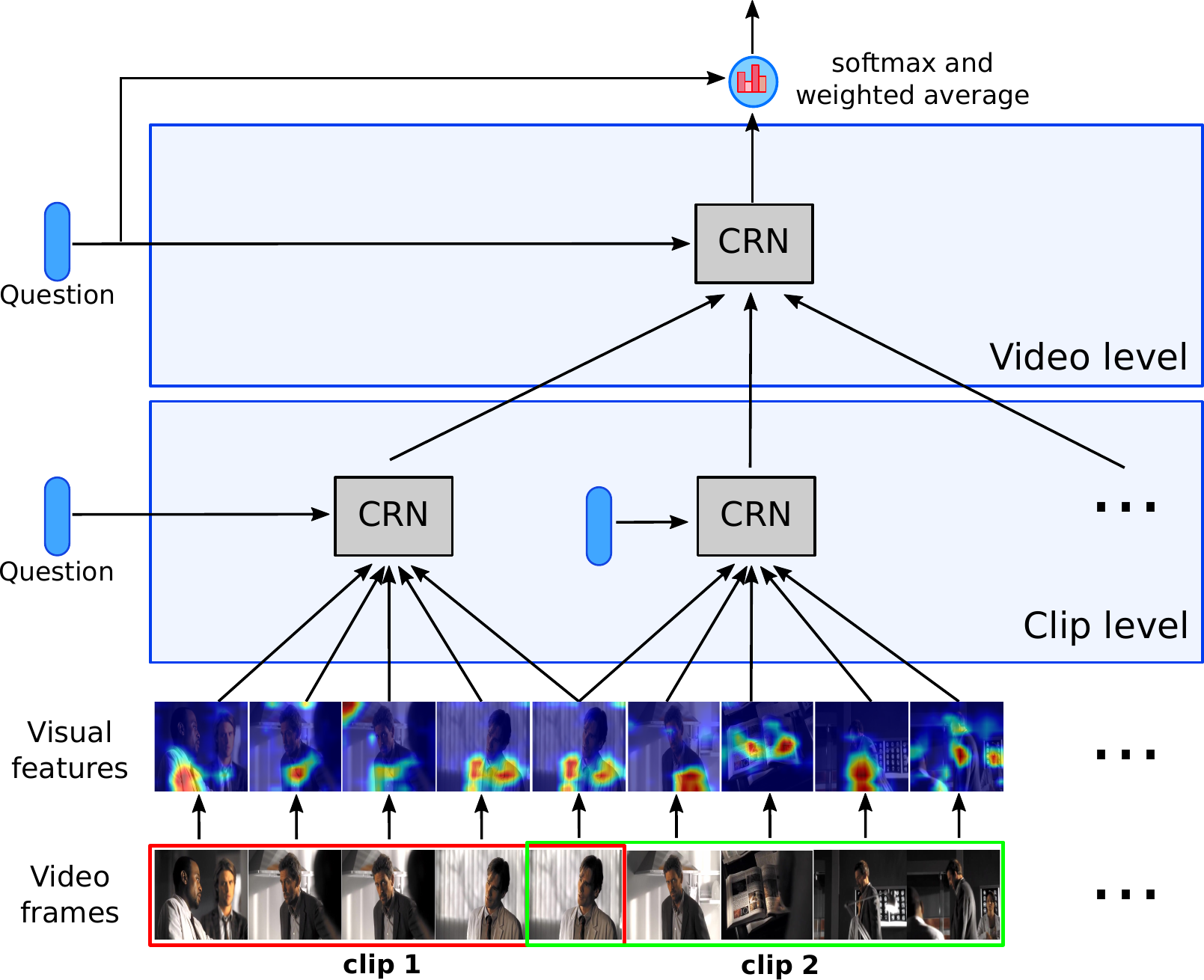}
\par\end{centering}
\caption{The adapted architecture of visual stream (Fig.\ref{fig:visual_stream})
for long-form video. At each level, only question-conditioned CRNs
are employed. The motion-conditioned CRNs are unnecessary as the low-level
motion features are less relevant in long-form media.\label{fig:visual_stream-long}}
\end{figure*}

All $\UnitName$ units at both levels take question representation
as conditioning features (See Fig. \ref{fig:visual_stream-long}).
Compared to the standard architecture introduced in Sec.~\ref{subsec:Visual-stream}
and Fig. \ref{fig:visual_stream}, we drop all $\UnitName$ units
that condition on motion features. This adaptation is to accommodate
the fact that low-level motion is less relevant than overall semantic
flow in both clip- and video-level. 

\subsection{Answer decoders\label{subsec:Answer-Decoders}}

Following the previous works \cite{fan2019heterogeneous,jang2017tgif,song2018explore},
we adopt different answer decoders depending on the task, as follows:
\begin{itemize}
\item QA pairs with open-ended answers are treated as multi-class classification
problems. For these, we employ a classifier which takes as input
the combination of the retrieved information from visual stream $\tilde{o}_{v}$,
the retrieved information from textual stream $\tilde{o}_{t}$ and
the question representation $q$, and computes answer probabilities
$p\in\mathbb{R}^{|\mathcal{A}|}$:
\begin{align}
y & =\text{ELU}\left(W_{o}\left[\tilde{o}_{v};\tilde{o}_{t};W_{q}q+b\right]+b\right),\label{eq:joint-embedding}\\
y^{\prime} & =\text{ELU}\left(W_{y}y+b\right),\label{eq:9}\\
p & =\textrm{softmax}\left(W_{y^{\prime}}y^{\prime}+b\right).
\end{align}
\item In the case of multi-choice questions where answer choices are available,
we iteratively treat each answer choice $\{a_{i}\}_{i=1,...,A}$ as
a query in exactly the manner that we handle the question. Eq.~\ref{eq:preselect_segment_level}
and Eq.~\ref{eq:preselection_subtitle}, therefore, take each of
the answer choices' representation $\{a_{i}\}$ as a conditioning
feature along with the question representation. Regarding Eq.~\ref{eq:joint-embedding},
it is replaced by:
\begin{equation}
y=\text{ELU}\left(W_{o}\left[\tilde{o}_{v_{\text{qa}}};\tilde{o}_{t_{\text{qa}}};W_{q}q+b;W_{a}a+b\right]+b\right),\label{eq:joint-embedding-mc}
\end{equation}
where $\tilde{o}_{v_{\text{qa}}}$ is output of visual stream respect
to queries as question and answer choices, whereas $\tilde{o}_{t_{\text{qa}}}$
is output of the textual stream counterpart.
\item For short-form $\Problem$, we simply drop the retrieved information
from textual stream $\tilde{o}_{t}$ in Eq.~\ref{eq:joint-embedding}
and Eq.~\ref{eq:joint-embedding-mc}. We also use the popular hinge
loss as what presents in \cite{jang2017tgif} for pairwise comparisons,
$\text{max}\left(0,1+s^{n}-s^{p}\right)$, between scores for incorrect
$s^{n}$ and correct answers $s^{p}$ to train the network:
\begin{equation}
s=W_{y^{\prime}}y^{\prime}+b.
\end{equation}
Regarding long-form $\Problem$, we use cross entropy as training
loss for fair comparisons with prior works.
\item For repetition count task, we use a linear regression function taking
$y^{\prime}$ in Eq.~\ref{eq:9} as input, followed by a rounding
function for integer count results. The loss for this task is Mean
Squared Error (MSE).
\end{itemize}

\subsection{Complexity analysis \label{subsec:Complexity-Analysis}}

We now provide a theoretical analysis of running time for CRN units
and HCRN. We will show that adding layers saves computation time,
just providing a justification for deep hierarchy.

\subsubsection{CRN units}

For clarity, let us recall the notations introduced in our CRN units:
$k_{\text{max}}$ is maximum subset (also tuple) size considered from
a given input array of $n$ objects, subject to $k_{\text{max}}<n$;
$t$ is number of size-$k$ subsets, ($k=2,3,...,k_{\text{max}}$),
randomly sampled from the input set; $g^{k}(.),h^{k}(.,.)$ and $p^{k}(.)$
are sub-networks for relation modeling, conditioning and aggregating,
respectively. In our implementation, $g^{k}(.)$ and $p^{k}(.)$ are
chosen to be set functions and $h^{k}(.,.)$ is a nonlinear transformation
that fuses modalities.

Each input object in CRN is arranged into a matrix of size $K\times F$,
where $K$ is the number of object elements and $F$ is the embedding
size for each element. All the operations involving input objects
are element-wise, that is, they are linear in time w.r.t. $K$. Assume
that the set function of order $k$ in the CRN's operation in Alg.
1 has linear time complexity in $k$. This holds true for most aggregation
functions such as mean, max, sum or product. With the relation orders
ranging from $k=2,3,...,k_{\text{max}}$ and sampling frequency $t$,
inference cost in time for a $\UnitName$ is:
\begin{equation}
\text{cost}_{\UnitName}\left(t,k_{\text{max}},K,F\right)=\text{cost}(g)+\text{cost}(h),\label{eq:CRN-complexity}
\end{equation}
where,
\begin{align*}
\text{cost}(g) & =\mathcal{O}\left(\frac{t}{2}k_{\text{max}}(k_{\text{max}}-1)KF\right),\\
\text{cost}(h) & =\mathcal{O}\left((4t+2)(k_{max}-1)KF^{2}\right).
\end{align*}
Here the running time of $\text{cost}(g)$ is quadratic in length
because each $g(.)$ that takes $k$ objects as input will cost $k$
time, for $k=2,3,...,k_{\text{max}}$. The running time of $\text{cost}(g)$
is quadratic in $F$ due to the feature transformation operation that
costs $F^{2}$ time. When $k_{\text{max}}\ll F$, $\text{cost}(h)$
will dominate the time complexity. However, since the function $h(.)$
involves matrix operations only, it is usually fast.

The unit produces an output array of length $k_{max}-1$, where each
output object is of the same size as the input objects.

\subsubsection{HCRN models}

We adhere to the complexity analysis of visual stream only which increases
linearly in complexity with a video length. The overall complexity
of $\ModelName$ depends on the design choice for each $\UnitName$
unit and the specific arrangement of $\UnitName$ units. For clarity,
let $t=2$ and $k_{\max}=n-1$, which are found to work well in experiments.
Let $L$ be the video length, organized into $N$ clips of length
$T$ each, i.e., $L=NT$.

\paragraph{2-level HCRN:}

Consider, for example, the 2-level architecture $\ModelName$, representing
clips and video. Each level is a stack of two $\UnitName$ layers,
one for motion conditioning followed by the other for linguistic conditioning.
The clip-level $\UnitName$s cost $N\times\text{cost}_{\UnitName}\left(2,T-1,1,F\right)$
time for motion conditioning and $N\times\text{cost}_{\UnitName}\left(2,T-3,1,F\right)$
time for question conditioning, where $\text{cost}_{\UnitName}$ is
the cost estimator in Eq.~(\ref{eq:CRN-complexity}). This adds to
roughly $\mathcal{O}\left(2TLF\right)+\mathcal{O}\left(10LF^{2}\right)$
time.

Now the output array of size $(T-4)\times F$ for the question-conditioned
clip-level $\UnitName$ becomes one in $N$ input objects the video-level
$\UnitName$s. The $\UnitName$s at the video level, therefore, take
a cost of $\text{cost}_{\UnitName}\left(2,N-1,T-4,F\right)$ time
for the motion-conditioned one and $\text{cost}_{\UnitName}\left(2,N-3,T-4,F\right)$
time for the question-conditioned one, respectively, totaling $\mathcal{O}\left(2NLF\right)+\mathcal{O}\left(10LF^{2}\right)$
in order. Here we have made use of the identity $L=NT$. The total
cost is therefore in the order of $\mathcal{O}\left(2(T+N)LF\right)+\mathcal{O}\left(20LF^{2}\right)$.

\paragraph{3-level HCRN:}

Let us now analyze a 3-level architecture $\ModelName$ that generalizes
the 2-level $\ModelName$. The $N$ clips are organized into $P$
sub-videos, each has $Q$ clips, i.e., $N=PQ$. Since the $\UnitName$s
at clip level remain the same, the first level costs $2TLF$ time
to compute as before. Moving to the next level, each sub-video $\UnitName$
takes as input an array of length $Q$, whose elements have size $(T-4)\times F$.
Applying the same logic as before, the set of sub-video-level CRNs
cost roughly $P\times\text{cost}_{\UnitName}\left(2,Q-1,T-4,F\right)$
time or approximately $\mathcal{O}\left(2\frac{N}{P}LF\right)+\mathcal{O}\left(10LF^{2}\right)$.
Here we have used the identities $N=PQ$ and $L=NT$.

A stack of two sub-video $\UnitName$s now produces an output array
of size $(Q-4)(T-4)\times F$, serving as an input object in an array
of length $P$ for the video-level $\UnitName$s. Thus the video-level
$\UnitName$s cost roughly 
\begin{eqnarray*}
\text{cost}_{\UnitName}\left(2,P-1,(Q-4)(T-4),F\right)
\end{eqnarray*}
time or approximately $\mathcal{O}\left(2PLF\right)+\mathcal{O}\left(10LF^{2}\right)$
. Here we again have used the identities $L=NT$ and $N=PQ$. Thus
the total cost is in the order of $\mathcal{O}\left(2(T+\frac{N}{P}+P)LF\right)+\mathcal{O}\left(30LF^{2}\right)$.

\subsubsection*{Deeper models might save time for long videos:}

Recall that the 2-level $\ModelName$ has time cost of 
\begin{eqnarray*}
\mathcal{O}\left(2(T+N)LF\right)+\mathcal{O}\left(20LF^{2}\right),
\end{eqnarray*}
and the 3-level HCRN the cost of 
\begin{eqnarray*}
\mathcal{O}\left(2(T+\frac{N}{P}+P)LF\right)+\mathcal{O}\left(30LF^{2}\right).
\end{eqnarray*}
Here the cost that is linear in $F$ is due to the $g$ functions,
and the quadratic cost in $F$ is due to the $h$ functions.

When going from 2-level to 3-level architectures, the cost for the
\emph{$g$ }functions\emph{ drops} by $\mathcal{O}\left(2(N-\frac{N}{P}-P)LF\right)$,
and the cost for the \emph{$h$} functions\emph{ increases} by $\mathcal{O}\left(10LF^{2}\right)$.
Now assuming $N\gg\max\left\{ P,\frac{N}{P}\right\} $, for example
$P\approx\sqrt{N}$ and the number of clips $N>20$. Then the linear
drop can be approximated further as $\mathcal{O}\left(2NLF\right)$.
As $N=\frac{L}{T}$, this can be written as $\mathcal{O}\left(2\frac{L^{2}}{T}F\right)$.
In practice the clip size $T$ is often fixed, thus the drop in the
$g$ functions scales quadratically with video length $L$, whereas
the increase in the $h$ functions scales linearly with $L$. This
suggests that \emph{going deeper in hierarchy could actually save
the running time for long videos}.

See Sec.~\ref{subsec:Deepening-model-hierarchy} for empirical validation
of the saving.

\section{Experiments \label{sec:Experiments}}

\subsection{Datasets}

We evaluate the effectiveness of the proposed $\UnitName$ unit and
the $\ModelName$ architecture on a series of short-form and long-form
$\Problem$ datasets. In particular, we use three different datasets
as benchmarks for the short-form $\Problem$, namely TGIF-QA \cite{jang2017tgif},
MSVD-QA \cite{xu2017video} and MSRVTT-QA \cite{xu2016msr}. All those
three datasets are collected from real-world videos. We also evaluate
HCRN on long-form $\Problem$ using one of the largest datasets publicly
available, TVQA \cite{lei2018tvqa}. Details of each benchmark are
as below.

\paragraph{TGIF-QA:}

This is currently the most prominent dataset for $\Problem$, containing
165K QA pairs and 72K animated GIFs. The dataset covers four tasks
addressing unique properties of video. Of which, the first three
require strong spatio-temporal reasoning abilities\emph{: Repetition
Count} - to retrieve the number of occurrences of an action, \emph{Repeating
Action}- multi-choice task to identify the action that is repeated
for a given number of times, \emph{State Transition} - multi-choice
tasks regarding temporal order of events. The last task - \emph{Frame
QA} - is akin to image QA where the answer to a given question can
be found from one particular video frame.

\paragraph{MSVD-QA:}

This is a small dataset of 50,505 question answer pairs annotated
from 1,970 short clips. Questions are of five types, including what,
who, how, when and where, of which 61\% of questions are used for
training whilst 13\% and 26\% are used as the validation set and test
set, respectively.

\paragraph{MSRVTT-QA:}

The dataset contains 10K videos and 243K question answer pairs. Similar
to MSVD-QA, questions are of five types. Splits for train, validation
and test are with the proportions are 65\%, 5\%, and 30\%, respectively.
Compared to the other two datasets, videos in MSRVTT-QA contain more
complex scenes. They are also much longer, ranging from 10 to 30 seconds
long, equivalent to 300 to 900 frames per video.

\paragraph{TVQA:}

This is one of the largest long-form Video QA datasets annotated from
6 different TV shows:\emph{ The Big Bang Theory}, \emph{How I Met
Your Mother}, \emph{Friends}, \emph{Grey's Anatomy}, \emph{House},
\emph{Castle}. There are total 152.5K question-answer pairs associated
with 5 answer choices each from 21,8K long clips of 60/90 secs which
comes down to 122K, 15,25K and 15,25K for train, validation and test
set, respectively. The dataset also provides start and end timestamps
for each question to limit the video portion where one can find corresponding
answers.

Regarding the evaluation metric, we mainly use accuracy in all experiments,
excluding those for repetition count on TGIF-QA dataset where Mean
Square Error (MSE) is applied.

\subsection{Implementation details}

\subsubsection{Feature extraction}

For short-form $\Problem$ datasets, each video is preprocessed into
$N$ short clips of fixed lengths, 16 frames each. In detail, we first
locate N equally spaced anchor frames. Each clip is then defined as
a sequence of 16 consecutive video frames taking a pre-computed anchor
as the middle frame. For the first and the last clip where frame indices
may exceed the boundaries, we repeat the first frame or the last frame
of the video multiple times until it fills up the clip\textquoteright s
size.

Given segmented clips, we extract motion features for each clip using
a pre-trained model of the ResNeXt-101\footnote{https://github.com/kenshohara/video-classification-3d-cnn-pytorch}
\cite{xie2017aggregated,hara2018can}. Regarding the appearance feature
used in the experiments, we take the \emph{pool5} output of ResNet
\cite{he2016deep} features as a feature representation of each frame.
This means we completely ignore the 2D structure of spatial information
of video frames which is likely to be beneficial for answering questions
particularly interested in object's appearance, such as those in the
Frame QA task in the TGIF-QA dataset. We are aware of this but deliberately
opt for light-weighted extracted features, and drive the main focus
of our work on the significance of temporal relation, motion, and
the hierarchy of video data by nature. Note that most of the videos
in the datasets in the short-form $\Problem$ category, except those
in the MSRVTT-QA dataset, are short. Hence, we intentionally divide
each video into 8 clips (8$\times$16 frames) to produce partially
overlapping frames between clips to avoid temporal discontinuity.
Longer videos in MSRVTT-QA are additionally segmented into 24 clips
of 16 frames each, primarily aiming at evaluating the model's ability
to handle very long sequences.

For the TVQA long-form dataset, we did a similar strategy as for short-video
divide each video into$N$ clips. However, as TVQA videos are longer
and only recorded at 3fps, we adapt by choosing $N=6$, each clip
contains 8 frames based on empirical experiences. The \emph{pool5}
output of ResNet features is also used as a feature representation
of each frame.

For the subtitles, the maximum subtitle's length is set at $256$.
We simply cut off those who are longer than that and do zero paddings
for those who are shorter. Subtitles are further divided into 6 segments
overlapping at half a segment size.

\subsubsection{Network training}

HCRN and its variations are implemented in Python 3.6 with Pytorch
1.2.0. Common settings include $d=512$, $t=2$ for both visual and
textual streams. For all experiments, we train the model using Adam
optimizer with a batch size of 32 , initially at a learning rate of
$10^{-4}$ and decay by half after every 5 epochs for counting task
in the TGIF-QA dataset and after every 10 epochs for the others. All
experiments are terminated after 25 epochs and reported results are
at the epoch giving the best validation accuracy. Depending on the
amount of training data and hierarchy depth, it may take around 4-30
hours of training on one single NVIDIA Tesla V100 GPU. Pytorch implementation
of the model is publicly available. \footnote{https://github.com/thaolmk54/hcrn-videoqa}.

As for experiments with the large-scale language representation model
BERT, we use the latest pre-trained model provide by Hugging Face\footnote{https://github.com/huggingface/transformers}.
We fine-tune the BERT model during training with a learning rate of
$2\times10^{-5}$ in all experiments.

\subsection{Quantitative results\label{subsec:Quantitative-results}}

We compare our proposed model with state-of-the-art methods (SOTAs)
on the aforementioned datasets. By default, we use pre-trained GloVe
embedding \cite{pennington2014glove} for word embedding following
by a BiLSTM for sequential modeling. Experiments using contextual
embeddings by a pre-trained BERT network \cite{devlin2018bert} is
explicitly specified with\emph{ ``(BERT)''}. Detailed experiments
for short-form $\Problem$ and long-form $\Problem$ are as the following.

\subsubsection{Short-form Video QA}

For TGIF-QA, we compare with most recent SOTAs, including \cite{fan2019heterogeneous,gao2018motion,jang2017tgif,li2019beyond},
over four tasks. These works, except for \cite{li2019beyond}, make
use of motion features extracted from either optical flow or 3D CNNs
and its variants.

\begin{figure}
\begin{centering}
\includegraphics[width=0.8\columnwidth,height=0.63\columnwidth]{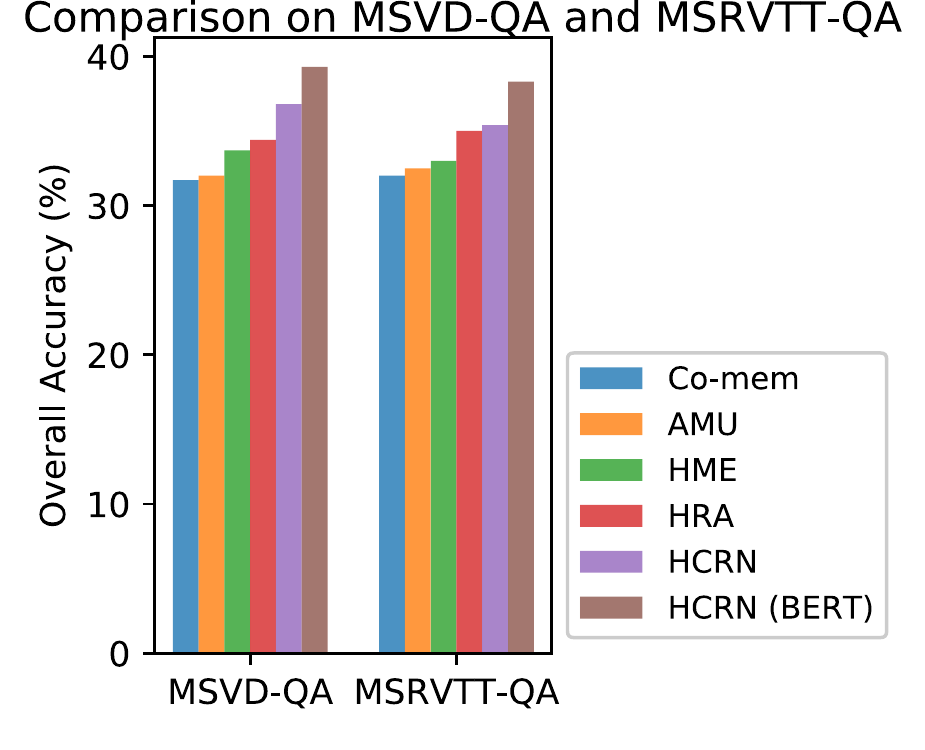}
\par\end{centering}
\caption{Performance comparison on MSVD-QA and MSRVTT-QA dataset with state-of-the-art
methods: Co-mem \cite{gao2018motion}, HME \cite{fan2019heterogeneous},
HRA \cite{chowdhury2018hierarchical}, and AMU \cite{xu2017video}.
\label{fig:msvd-msrvtt}}
\medskip{}
\end{figure}

The results are summarized in Table~\ref{tab:tgif} for TGIF-QA,
and in Fig.~\ref{fig:msvd-msrvtt} for MSVD-QA and MSRVTT-QA. Results
of the previous works are taken from either the original papers or
what reported by \cite{fan2019heterogeneous}. It is clear that our
model consistently outperforms or is competitive with SOTA models
on all tasks across all datasets. The improvements are particularly
noticeable when strong temporal reasoning is required, i.e., for the
questions involving actions and transitions in TGIF-QA. These results
confirm the significance of considering both near-term and far-term
temporal relations toward finding correct answers. In addition, results
on the TGIF-QA dataset over 10 runs of different random seeds confirm
the robustness of the CRN units against the randomness in input subsets
selection. More analysis on how relations affect the model's performance
is provided in later ablation studies.

Regarding results with the recent advance in language representation
model BERT, it does not have much effect on the results across tasks
requiring strong temporal reasoning in the TGIF-QA. This can be explained
by the fact that questions in this dataset are relatively short and
all questions are created from a limited number of patterns, hence,
contextual embeddings do not account much benefit. Whereas the Frame
QA task relies on much richer vocabulary, thanks to its free-form
questions, hence, contextual embeddings extracted by BERT greatly
boost the model's performance on this task. We also empirically find
out that fine-tune only the last two layers of the BERT network gives
more favorable performance comparing to fine-tuning all layers.

The MSVD-QA and MSRVTT-QA datasets represent highly challenging benchmarks
for machine compared to the TGIF-QA, thanks to their open-ended nature.
Our model $\ModelName$ outperforms existing methods on both datasets,
achieving 36.8\% and 35.4\% accuracy which are 2.4 points and 0.4
points improvement on MSVD-QA and MSRVTT-QA, respectively. This suggests
that the model can handle both small and large datasets better than
existing methods. We also fine-tune the contextual embeddings by
BERT on these two datasets. Results show that the contextual embeddings
bring great benefits on both MSVD-QA and MSRVTT-QA, achieving 39.3\%
and 38.3\% accuracy, respectively. These figures are consistent with
the result on the Frame QA task of the TGIF-QA dataset. Please note
that we also fine-tune only last two layers of the BERT network in
these experiments simply due to empirical results.

Finally, we provide a justification for the competitive performance
of our $\ModelName$ against existing rivals by comparing model features
in Table~\ref{tab:Model-design-choices}. Whilst it is not straightforward
to compare head-to-head on internal model designs, it is evident that
effective video modeling necessitates handling of motion, temporal
relation and hierarchy at the same time. We will back this hypothesis
by further detailed studies in Sec.~\ref{subsec:Ablation-Studies}
(for motion, temporal relations, shallow hierarchy) and Sec.~\ref{subsec:Deepening-model-hierarchy}
(deep hierarchy).

\subsubsection{Long-form Video QA}

For TVQA, we compare our method to several baselines as well as state-of-the-art
methods (See Table~\ref{tab:tvqa}). The dataset is relatively new
and due to the challenges with long-form $\Problem$, there are only
several attempts in benchmarking it. Comparisons of the proposed method
with the baselines are made on the validation set while results of
compared methods are taken as reported in original publications. Experiments
using timestamp information are indicated by \emph{w/ ts} and those
with full-length subtitles are with \emph{w/o ts. }If not indicated
explicitly, ``V.'', ``S.'', ``Q.'' short for visual features,
subtitle features and question features, respectively. The evaluated
settings are as follows:
\begin{itemize}
\item \emph{(B) Q. }: This is the simplest baseline where we only use question
features for predicting the correct answer. Results in this setting
will reveal how much our model relying on the linguistic bias to arrive
at correct answers.
\item \emph{(B) S. + Q. (w/o pre-selection)}: In this baseline, we use question
and subtitle features for prediction. Both question and subtitle features
are simply obtained by concatenating the final output hidden states
of forward and backward LSTM passes and further pass to a classifier
as in Sec.~\ref{subsec:Answer-Decoders}.
\item \emph{(B) S. + Q. (w/ pre-selection)}: This baseline is to show the
significance of pre-selection in textual stream as in Eq.~\ref{eq:preselection_subtitle}.
Having said that, question features and selective output between the
question and the subtitles as explained in Eq.~\ref{eq:preselection_subtitle}
are fed into a classifier for prediction.
\item \emph{(B) V. + Q.}: In this baseline, we simply apply average pooling
to smash visual features of entire videos into a vector and further
combine it with question features for prediction.
\item \emph{(B) S. + V. + Q. (w/ pre-selection)}: We combine two baselines
\emph{``(B) S. + Q. (w/ pre-selection)''} and \emph{``(B) V. +
Q.''} right above. Given that, subtitle features are extracted with
pre-selection as in Eq.~\ref{eq:preselection_subtitle} and video
features are smashed over space-time before going through a classifier.
\item \emph{(HCRN) S. + Q.}: This is to evaluate the effect of the textual
stream alone in our proposed network architecture as described in
Fig.~\ref{fig:textual_stream}.
\item \emph{(HCRN) V. + Q.}: This is to evaluate the effect of the visual
stream counterpart alone in Fig.~\ref{fig:visual_stream-long}.
\item \emph{(HCRN) S. + V. + Q.}: Our full proposed architecture with the
presence of both two modalities.
\end{itemize}
As shown in Table~\ref{tab:tvqa}, using BERT for contextual word
embeddings significantly improves performance comparing to GloVe embeddings
and BiLSTM counterpart. The results are consistent with what reported
by \cite{yang2020bert}. Although our model only achieves competitive
performance with \cite{yang2020bert} in the setting of using timestamps
(71.3 vs. 72.1), we significantly outperform them by 3.0 absolute
points on the more challenging setting when we do not have access
to timestamp indicators of where answers located. This clearly shows
that our bottom-up approach is promising to solve long-form Video
QA in its general setting. 

Even though BERT is designed to do contextual embedding which pre-computes
some word relations, HCRN still shows additional benefit in modeling
relations between segments in the passage. However, this benefit is
not as clearly demonstrated as in the case of using GloVe embedding
coupled with BiLSTM (Table~\ref{tab:tvqa} - Exp. 9 vs. Exp. 7 and
Exp. 13 vs. Exp. 14). To deeper understanding the behavior of HCRN,
we will concentrate our further analysis using comparison with GloVe
+ LSTM.

It clearly shows that using output hidden states of BiLSTM to represent
subtitle features directly for classification has a very limited effect
(See Table~\ref{tab:tvqa} - Exp. 3 vs Exp. 4). The results could
be explained as LSTM fails to handle such long subtitles of hundreds
of words. In the meantime, pre-selection plays a critical role to
find relevant information in the subtitles to a question which boosts
performance from 42.3\% to 57.9\% when using full subtitles and from
41.8\% to 62.1\% when using timestamps (See Exp. 5). Our $\ModelName$
further improves approximately 1.0 points (without timestamp annotation)
and 1.5 points (with timestamp annotation) comparing to the baseline
of GloVe embeddings and BiLSTM with pre-selection for the textual
stream only setting. As for the effects on visual stream, our $\ModelName$
gains around 1.0 points over the baseline of averaging visual features.
This leads to an improvement of north of 1.0 points when leveraging
both visual and textual stream together (See Exp. 12 vs. Exp. 15)
on both settings with timestamp annotation and without timestamp annotation.
Even though our results are behind \cite{lei2018tvqa}, we wish to
point out that their context matching model with the bi-directional
attention flow (BiDAF) \cite{seo2016bidirectional} significantly
contributes to the performance. Whereas, our model only makes use
of vanilla BiLSTM to sequence modeling. Therefore, a direct comparison
between the two methods is unfair. We instead mainly compare our method
with a simple variant of \cite{lei2018tvqa} by dropping off the BiDAF
to show the contribution of our CRN unit as well as the HCRN network.
This is equivalent to the baseline \emph{(B) S. + V. + Q. (w/ pre-selection)
}in this paper.

\begin{table*}
\begin{centering}
\begin{tabular}{l|c|c|c|c}
\hline 
Model & Action & Trans. & Frame & Count\tabularnewline
\hline 
ST-TP \cite{jang2017tgif} & 62.9 & 69.4 & 49.5 & 4.32\tabularnewline
Co-mem \cite{gao2018motion} & 68.2 & 74.3 & 51.5 & 4.10\tabularnewline
PSAC \cite{li2019beyond} & 70.4 & 76.9 & 55.7 & 4.27\tabularnewline
HME \cite{fan2019heterogeneous} & 73.9 & 77.8 & 53.8 & 4.02\tabularnewline
\hline 
HCRN{*} & \textbf{75.2}$\pm$0.4 & \textbf{81.3}$\pm$0.2 & 55.9$\pm$0.3 & \textbf{3.93}$\pm$0.03\tabularnewline
HCRN (embeddings with BERT) & 69.8 & 79.8 & \textbf{57.9} & 3.96\tabularnewline
\hline 
\end{tabular}\medskip{}
\par\end{centering}
\caption{Comparison with the state-of-the-art methods on TGIF-QA dataset. For
count, the lower the better (MSE) and the higher the better for the
others (accuracy). {*}Means with standard deviations over 10 runs.\label{tab:tgif}}
\end{table*}

\begin{table}
\begin{centering}
{\scriptsize{}}%
\begin{tabular}{l|c|c|c|c}
\hline 
Model & Appear. & Motion & Hiera. & Relation\tabularnewline
\hline 
ST-TP \cite{jang2017tgif} & $\checkmark$ & $\checkmark$ &  & \tabularnewline
Co-mem \cite{gao2018motion} & $\checkmark$ & $\checkmark$ &  & \tabularnewline
PSAC \cite{li2019beyond} & $\checkmark$ &  &  & \tabularnewline
HME \cite{fan2019heterogeneous} & $\checkmark$ & $\checkmark$ &  & \tabularnewline
$\ModelName$ & $\checkmark$ & $\checkmark$ & $\checkmark$ & $\checkmark$\tabularnewline
\hline 
\end{tabular}\medskip{}
\par\end{centering}
\caption{Model design choices and input modalities in comparison. See Table~\ref{tab:tgif}
for corresponding performance on TGIF-QA dataset. \label{tab:Model-design-choices}}
\medskip{}
\end{table}

\begin{table*}
\begin{centering}
\begin{tabular}{l|l|c|c}
\hline 
\multirow{2}{*}{Exp.} & \multirow{2}{*}{Model} & \multicolumn{2}{c}{Val. Acc.}\tabularnewline
\cline{3-4} \cline{4-4} 
 &  & w/o ts & w/ ts\tabularnewline
\hline 
 & \textbf{State-of-the-art methods} &  & \tabularnewline
1 & $\quad$TVQA S.+Q.+V. \cite{lei2018tvqa}\cite{yang2020bert} & 64.7 & 67.7\tabularnewline
2 & $\quad$VideoQABERT \cite{yang2020bert} & 63.1 & \textbf{72.1}\tabularnewline
\hline 
 & \textbf{Textual stream only} &  & \tabularnewline
3 & $\quad$(B) Q. & 41.6 & 41.6\tabularnewline
4 & $\quad$(B) S. + Q. (w/o pre-selection) & 42.3 & 41.8\tabularnewline
5 & $\quad$(B) S. + Q. (w/ pre-selection) & 57.9 & 62.1\tabularnewline
6 & $\quad$(B) Q. (BERT) & 44.2 & 44.2\tabularnewline
7 & $\quad$(B) S. + Q. (BERT) & 65.1 & 71.1\tabularnewline
8 & $\quad$(HCRN) S. + Q. & 59.0 & 63.6\tabularnewline
9 & $\quad$(HCRN) S. + Q. (BERT) & 66.0 & 71.1\tabularnewline
\hline 
 & \textbf{Visual stream only} &  & \tabularnewline
10 & $\quad$(B) V. + Q. & 42.2 & 42.2\tabularnewline
11 & $\quad$(HCRN) V. + Q. & 43.1 & 43.1\tabularnewline
\hline 
 & \textbf{Two streams} &  & \tabularnewline
12 & $\quad$(B) S. + V. + Q. (w/ pre-selection) & 58.5 & 63.2\tabularnewline
13 & $\quad$(B) S. + V. + Q. (BERT) & 65.7 & 71.4\tabularnewline
14 & $\quad$(HCRN) S. + V. + Q. (BERT) & \textbf{66.1} & 71.3\tabularnewline
15 & $\quad$(HCRN) S. + V. + Q. & 59.1 & 64.7\tabularnewline
\hline 
\end{tabular}\medskip{}
\par\end{centering}
\caption{Comparison with baselines and state-of-the-art methods on TVQA dataset.
\emph{w/o ts}: without using timestamp annotation to limit the search
space of where to find answers; \emph{w/ ts}: making use of the timestamp
annotation.\label{tab:tvqa}}
\end{table*}

\subsection{Ablation studies \label{subsec:Ablation-Studies}}

\begin{table*}
\centering{}%
\begin{tabular}{l|c|c|c|c}
\hline 
Model & Action & Transition & FrameQA & Count\tabularnewline
\hline 
\hline 
\textbf{Relations $(k_{max},t)$} &  &  &  & \tabularnewline
$\quad$$k_{max}=1,t=1$ & 72.8 & 80.7 & 56.3 & 3.87\tabularnewline
$\quad$$k_{max}=1,t=3$ & 73.4 & 81.1 & 56.0 & 3.94\tabularnewline
$\quad$$k_{max}=1,t=5$ & 74.0 & 80.3 & 56.4 & 3.86\tabularnewline
$\quad$$k_{max}=1,t=7$ & 74.8 & 80.0 & 56.5 & 3.85\tabularnewline
$\quad$$k_{max}=1,t=9$ & 73.7 & 81.1 & 56.4 & 3.82\tabularnewline
$\quad$$k_{max}=2,t=2$ & 72.8 & 80.8 & 56.3 & 3.80\tabularnewline
$\quad$$k_{max}=2,t=9$ & 73.1 & 81.4 & 56.0 & 3.85\tabularnewline
$\quad$$k_{max}=4,t=2$ & 74.2 & 81.5 & 56.8 & 3.88\tabularnewline
$\quad$$k_{max}=4,t=9$ & 73.4 & 81.8 & 56.5 & 3.83\tabularnewline
$\quad$$k_{max}=\left\lfloor n/2\right\rfloor ,t=2$ & 74.7 & 81.1 & 55.7 & 3.85\tabularnewline
$\quad$$k_{max}=\left\lfloor n/2\right\rfloor ,t=9$ & 74.7 & 81.0 & 55.4 & 3.94\tabularnewline
$\quad$$k_{max}=n-1,t=1$ & 74.2 & 80.8 & 55.3 & 3.98\tabularnewline
$\quad$$k_{max}=n-1,t=3$ & 75.2 & 81.1 & 56.2 & 4.00\tabularnewline
$\quad$$k_{max}=n-1,t=5$ & 75.0 & 81.3 & 55.6 & 3.95\tabularnewline
$\quad$$k_{max}=n-1,t=7$ & 75.3 & 81.9 & 55.8 & 4.00\tabularnewline
$\quad$$k_{max}=n-1,t=9$ & 74.9 & 81.1 & 55.6 & 3.94\tabularnewline
$\quad$Fix $k=k_{max},k_{max}=n-1,t=2$ & 72.9 & 80.2 & 56.5 & 3.90\tabularnewline
\hline 
\textbf{Hierarchy} &  &  &  & \tabularnewline
$\quad$$1$-level, video $\UnitName$ only & 72.7 & 81.4 & 57.2 & 3.88\tabularnewline
$\quad$$1.5$-level, clips$\rightarrow$pool & 73.1 & 81.2 & 57.2 & 3.88\tabularnewline
\hline 
\textbf{Motion conditioning} &  &  &  & \tabularnewline
$\quad$w/o motion & 69.8 & 78.4 & 57.9 & 4.38\tabularnewline
$\quad$w/o short-term motion & 74.1 & 80.9 & 56.4 & 3.87\tabularnewline
$\quad$w/o long-term motion & 75.8 & 80.8 & 56.8 & 3.97\tabularnewline
\hline 
\textbf{Linguistic conditioning} &  &  &  & \tabularnewline
$\quad$w/o linguistic condition & 68.7 & 80.5 & 56.6 & 3.92\tabularnewline
$\quad$w/o question @clip level & 75.0 & 81.0 & 56.0 & 3.85\tabularnewline
$\quad$w/o question @video level & 74.2 & 81.0 & 55.3 & 3.90\tabularnewline
\hline 
\textbf{Multiplicative relation} &  &  &  & \tabularnewline
$\quad$w/o MUL. in all CRNs & 74.0 & 81.7 & 55.8 & 3.86\tabularnewline
$\quad$w/ MUL. for both question \& motion & 75.4 & 80.2 & 55.1 & 3.98\tabularnewline
\hline 
\textbf{Subset sampling} &  &  &  & \tabularnewline
$\quad$Sampled from pre-computed superset & 75.2 & 81.3 & 55.9 & 3.90\tabularnewline
$\quad$Directly sampled & 75.3 & 81.8 & 55.3 & 3.89\tabularnewline
\hline 
\end{tabular}\medskip{}
\caption{Ablation studies on TGIF-QA dataset. For count, the lower the better
(MSE) and the higher the better for the others (accuracy). When not
explicitly specified, we use $k_{max}=n-1,t=2$ for relation order
and sampling resolution. \label{tab:Ablation-tgif}}
\medskip{}
\end{table*}

\begin{table*}
\begin{centering}
\begin{tabular}{l|l|c|c}
\hline 
\multirow{2}{*}{Exp.} & \multirow{2}{*}{Model} & \multicolumn{2}{c}{Val. Acc.}\tabularnewline
\cline{3-4} \cline{4-4} 
 &  & w/o ts & w/ ts\tabularnewline
\hline 
 & \textbf{Textual stream only} &  & \tabularnewline
1 & S. + Q. (w/ MUL. + w/ LSTM) & 59.0 & 63.6\tabularnewline
2 & S. + Q. (w/ MUL. + w/o LSTM) & 57.1 & 63.6\tabularnewline
3 & S. + Q. (w/o MUL. + w/ LSTM) & 59.1 & 63.9\tabularnewline
4 & S. + Q. (w/o MUL. + w/o LSTM) & 57.4 & 63.5\tabularnewline
\hline 
 & \textbf{Visual stream only} &  & \tabularnewline
5 & V. + Q. (w/ MUL. + w/ LSTM) & 42.7 & 42.7\tabularnewline
6 & V. + Q. (w/ MUL. + w/o LSTM) & 43.1 & 43.1\tabularnewline
7 & V. + Q. (w/o MUL. + w/ LSTM) & 42.2 & 42.2\tabularnewline
8 & V. + Q. (w/o MUL. + w/o LSTM) & 42.1 & 42.1\tabularnewline
\hline 
 & \textbf{Two streams} &  & \tabularnewline
9 & S. + V. + Q. (S. w/ MUL. + LSTM, V. w/ MUL.) & 59.1 & 64.7\tabularnewline
\hline 
\end{tabular}\medskip{}
\par\end{centering}
\caption{Ablation studies on TVQA dataset.\label{tab:tvqa-ablation}}
\medskip{}
\end{table*}

To provide more insights about the roles of $\ModelName$'s components,
we conduct extensive ablation studies on the TGIF-QA and TVQA dataset
with a wide range of configurations.\emph{ }The results are reported
in Table~\ref{tab:Ablation-tgif} for the short-form $\Problem$
and in Table\ \ref{tab:tvqa-ablation} for the long-form $\Problem$. 

\subsubsection{Short-form Video QA}

Overall we find that our sampling method does not hurt the $\ModelName$
performance that much while it is much more effective than the one
used by \cite{zhou2018temporal} in terms of complexity. We also find
that ablating any of the design components or $\UnitName$ units would
degrade the performance for temporal reasoning tasks (actions, transition
and action counting). The effects are detailed as follows.

\paragraph{Effect of relation order $k_{max}$ and resolution $t$:}

Without relations ($k_{max}=1,t=1$) the performance suffers, specifically
on actions and events reasoning whereas counting tends to be better.
This is expected since those questions often require putting actions
and events in relation with a larger context (e.g., what happens before
something else) while motion flow is critical for counting but for
far-term relations. In this case, most of the tasks benefit from increasing
sampling resolution $t$ $(t>1)$ because of better chance to find
a relevant frame as well as the benefits of the far-term temporal
relation learned by the aggregating sub-network $p^{k}(.)$ of the
$\UnitName$ unit. However, when taking relations into account ($k_{max}>1$),
we find that $\ModelName$ is robust against sampling resolution
$t$ but depends critically on the maximum relation order $k_{max}$.
The relative independence w.r.t. $t$ can be due to visual redundancy
between frames, so that resampling may capture almost the same information.
On the other hand, when considering only low-order object relations,
the performance is significantly dropped in action and transition
tasks while it is slightly better for counting and frame QA. These
results confirm that high-order relations are required for temporal
reasoning. As the frame QA task requires only reasoning on a single
frame, incorporating temporal information might confuse the model.
Similarly, when the model only makes use of the high-order relations
(\emph{Fix} $k=k_{max},k_{max}=n-1,t=2$), the performance suffers,
suggesting combining both low-order object relations and high-order
object relations is a lot more efficient.

\paragraph{Effect of hierarchy:}

We design two simpler models with only one $\UnitName$ layer: 
\begin{itemize}
\item $1$\emph{-level, $1$ $\UnitName$ video on key frames only}: Using
only one $\UnitName$ at the video-level whose input array consists
of key frames of the clips. Note that video-level motion features
are still maintained. 
\item $1.5$-\emph{level, clip $\UnitName$s $\rightarrow$} \emph{pooling}:
Only the clip-level $\UnitName$s are used, and their outputs are
mean-pooled to represent a given video. The pooling operation represents
a simplistic relational operation across clips. The results confirm
that a hierarchy is needed for high performance on temporal reasoning
tasks.
\end{itemize}

\paragraph{Effect of motion conditioning:}

We evaluate the following settings: 
\begin{itemize}
\item \emph{w/o short-term motions}: Remove all $\UnitName$ units that
condition on the short-term motion features (clip level) in the $\ModelName$.
\item \emph{w/o long-term motions}: Remove the $\UnitName$ unit that conditions
on the long-term motion features (video level) in the $\ModelName$.\emph{ }
\item \emph{w/o motions}: Remove motion feature from being used by $\ModelName$.
We find that motion, in agreeing with prior arts, is critical to detect
actions, hence computing action count. Long-term motion is particularly
significant for the counting task, as this task requires maintaining
a global temporal context during the entire process. For other tasks,
short-term motion is usually sufficient. E.g. in the action task,
wherein one action is repeatedly performed during the entire video,
long-term context contributes little. Not surprisingly, motion does
not play a positive role in answering questions on single frames as
only appearance information needed.
\end{itemize}

\paragraph{Effect of linguistic conditioning and multiplicative relation:}

Linguistic cues represent a crucial context for selecting relevant
visual artifacts. For that we test the following ablations:
\begin{itemize}
\item \emph{w/o question @clip level}: Remove the $\UnitName$ unit that
conditions on question representation at clip level. 
\item \emph{w/o question @video level}: Remove the $\UnitName$ unit that
conditions on question representation at video level.
\item \emph{w/o linguistic condition: }Exclude all $\UnitName$ units conditioning
on linguistic cue while the linguistic cue is still in the answer
decoder. Likewise, the multiplicative relation form offers a selection
mechanism. Thus we study its effect as follows:
\item \emph{w/o MUL.} \emph{in all CRNs}: Exclude the use of multiplicative
relations in all $\UnitName$ units.
\item \emph{w/ MUL. relation for question \& motion}: Leverage multiplicative
relations in all $\UnitName$ units. 
\end{itemize}
We find that the conditioning question provides an important context
for encoding video. Conditioning features (motion and language), through
the multiplicative relation as in Eq.~\ref{eq:conditioning_mul},
offers further performance gain in all tasks rather than Frame QA,
possibly by selectively passing question-relevant information up the
inference chain.

\paragraph{Effect of subset sampling}

We conduct experiments with the full model of Fig.~\ref{fig:visual_stream}
(a) with $k_{max}=n-1,t=2$. The experiment \emph{``Sampled from
pre-computed superset''} refers to our $\UnitName$ units of using
the same sampling trick as what is in \cite{zhou2018temporal}, where
the set $Q^{k}$ is sampled from a pre-computed collection of all
possible size-$k$ subsets. \emph{``Directly sampled''}, in contrast,
refers to our sampling method as described in Alg. 1. The empirical
results show that directly sampling size-$k$ subsets from an input
set does not degrade much of the performance of the $\ModelName$
for short-form $\Problem$, suggesting it a better choice when dealing
with large input sets in size to reduce the complexity of the $\UnitName$
units.

\subsubsection{Long-form Video QA}

We focus on studying the effect of different options for the conditioning
sub-network $h^{k}(.,.)$ in Sec.~\ref{subsec:Relation-Network}
which reflects the flexibility of our model when dealing with different
forms of input modalities. The options are:
\begin{itemize}
\item \emph{S. + Q. (w/ MUL. + w/ LSTM)}: Only textual stream is used.
The conditioning sub-network $h^{k}(.,.)$ is with multiplicative
relation between tuples of segments and conditioning feature, and
coupled with BiLSTM as formulated in Eqs.~(\ref{eq:conditioning},
\ref{eq:conditioning_bilstm} and \ref{eq:max_temporally}).
\item \emph{S. + Q. (w/ MUL. + w/o LSTM)}: Remove the BiLSTM network in
the experiment \emph{S. + Q. (w/ MUL. + w/ LSTM)} to evaluate the
effect of sequential modeling in textual stream.
\item \emph{S. + Q. (w/o MUL. + w/ LSTM)}: The multiplicative relation in
the experiment \emph{S. + Q. (w/ MUL. + w/ LSTM)} is now replaced
with a simple concatenation of conditioning feature and tuples of
segments.
\item \emph{S. + Q. (w/o MUL. + w/o LSTM)}: Remove the BiLSTM network in
the right above experiment\emph{ S. + Q. (w/o MUL. + w/ LSTM)} to
evaluate the effect of when both selective relation and sequential
modeling are missing.
\item \emph{V. + Q. (w/ MUL. + w/ LSTM)}: Only visual stream is under consideration.
We use the multiplicative form as in Eq.~\ref{eq:conditioning_mul}
for the conditioning sub-network instead of a simple concatenation
operation in Eq.~\ref{eq:conditioning_concat}. We additionally use
a BiLSTM network for sequential modeling as same as the way we have
done with the textual stream. This is because both visual content
and textual content are temporal sequences by nature.
\item \emph{V. + Q. (w/ MUL. + w/o LSTM)}: Similar to the above experiment
\emph{V. + Q. (w/ MUL. + w/ LSTM)} but without the use of the BiLSTM
network.
\item \emph{V. + Q. (w/o MUL. + w/ LSTM)}: The conditioning sub-network
is simply a tensor concatenation operation. This is to compare against
the one using multiplicative form \emph{V. + Q. (w/ MUL. + w/ LSTM)}.
\item \emph{V. + Q. (w/o MUL. + w/o LSTM)}: Similar to \emph{V. + Q. (w/o
MUL. + w/ LSTM)} but without the use of the BiLSTM network for sequential
modeling.
\item \emph{S. + V. + Q. (S. w/ MUL. + LSTM, V. w/ MUL.)}: Both two streams
are present for prediction. We combine the best option of each network
stream, \emph{S. + Q. (w/ MUL. + w/ LSTM)} for the textual stream
and \emph{V. + Q. (w/ MUL. + w/o LSTM) }for the visual stream for
comparison. 
\end{itemize}
It is empirically shown that the simple concatenation as in Eq.~\ref{eq:conditioning_concat}
is insufficient to combine conditioning features and output of relation
network in this dataset. In the meantime, the multiplicative relation
between the conditioning features and those relations is a better
fit as it greatly improves the model's performance, especially in
the visual stream. This shows the consistency in empirical results
between long-form $\Problem$ and shot-form $\Problem$ where the
relation between visual content and the query is more about multiplicative
(selection) rather than simple additive. 

On the other side, sequential modeling with BiLSTM is more favorable
for textual stream than visual stream even though both streams are
sequential by nature. This well aligns with our analysis in Sec.~\ref{subsec:Relation-Network}.

\begin{table}
\begin{centering}
\begin{tabular}{l|c|c}
\hline 
\multirow{2}{*}{Depth of hierarchy} & \multicolumn{2}{c}{Overall Acc.}\tabularnewline
\cline{2-3} \cline{3-3} 
 & Val. & Test\tabularnewline
\hline 
$2$-level, $24$ clips $\rightarrow$ $1$ vid & 35.4 & 35.5\tabularnewline
$3$-level, $24$ clips $\rightarrow$ $4$ sub-vids $\rightarrow$
$1$ vid & 35.1 & 35.4\tabularnewline
\hline 
\end{tabular}\medskip{}
\par\end{centering}
\caption{Results for going deeper hierarchy on MSRVTT-QA dataset. Run time
is reduced by factor of $4$ for going from 2-level to 3-level hierarchy.
\label{tab:scale_analysis-msrvtt}}
\medskip{}
\end{table}

\subsubsection{Deepening model hierarchy saves time \label{subsec:Deepening-model-hierarchy}}

We test the scalability of the HCRN on long videos in the MSRVTT-QA
dataset, which are organized into 24 clips (3 times longer than the
other two datasets). We consider two settings:
\begin{itemize}
\item \emph{$2$-level hierarchy,} \emph{$24$ clips}$\rightarrow$$1$
\emph{vid}: The model is as illustrated in Fig.~\ref{fig:visual_stream},
where 24 clip-level $\UnitName$s are followed by a video-level $\UnitName$.
\item \emph{$3$-level hierarchy}, $24$ \emph{clips}$\rightarrow$$4$\emph{
sub-vid}s$\rightarrow$$1$ \emph{vid}: Starting from the 24 clips
as in the $2$-level hierarchy, we group 24 clips into 4 sub-videos,
each is a group of 6 consecutive clips, resulting in a $3$-level
hierarchy. These two models are designed to have a similar number
of parameters, approx. 44M.
\end{itemize}
The results are reported in Table~\ref{tab:scale_analysis-msrvtt}.
Unlike existing methods which usually struggle with handling long
videos, our method is scalable for them by offering deeper hierarchy,
as analyzed theoretically in Sec.~\ref{subsec:Complexity-Analysis}.
The theory suggests that using a deeper hierarchy can reduce the training
time and inference time for $\ModelName$ when the video is long.
This is validated in our experiments, where we achieve \emph{4 times
reduction in training and inference time} by going from 2-level $\ModelName$
to 3-level counterpart whilst maintaining competitive performance.

\section{Discussion \label{sec:Discussion}}

HCRN presents a new class of neural architectures for multimodal Video
QA, pursuing the ease of model construction through reusable uniform
building blocks. Different from temporal attention based approaches
which put effort into selecting objects, HCRN concentrates on modeling
relations and hierarchy in different input modalities in $\Problem$.
In the scope of this work, we study how to deal with visual content
and subtitles where applicable. The visual and text streams share
similar structure in terms of near-term, far-term relations and information
hierarchy. The difference in methodology and design choices between
ours and the existing works leads to distinctive benefits in different
scenarios as empirically proven. 

Within the scope of this paper we did not consider the audio channel
and early-fusion between modalities, leaving these open for future
work. Since the CRN is generic, we envision a HCRN for the audio stream
similar to the visual stream. The modalities can be combined at any
hierarchical level, or any step within the same level. For example,
as the audio, subtitle and visual content are partly synchronized,
we can use a sub-HCRN to represent the three streams per segment.
Alternatively, we can fuse the modalities into the same CRN as long
as the feature representations are projected onto the same tensor
space. Future works will also include object-oriented representation
of videos as these are native to CRNs, thanks to its generality. As
CRN is a relational model, both cross-object relations and cross-time
relations can be modeled. These are likely to improve the interpretability
of the model and get closer to how human reasons across multiple modalities.
CRN units can be further augmented with attention mechanisms to cover
better object selection ability, so that related tasks such as frame
QA in the TGIF-QA dataset can be further improved.

\subsection{Conclusion}

We introduced a general-purpose neural unit called Conditional Relational
Networks (CRNs) and a method to construct hierarchical networks for
multimodal $\Problem$ using $\UnitName$s as building blocks. A CRN
is a relational transformer that encapsulates and maps an array of
tensorial objects into a new array of relations, all conditioned on
a contextual feature. In the process, high-order relations among input
objects are encoded and modulated by the conditioning feature. This
design allows flexible construction of sophisticated structure such
as stack and hierarchy, and supports iterative reasoning, making it
suitable for QA over multimodal and structured domains like video.
The $\ModelName$ was evaluated on multiple $\Problem$ datasets covering
both short-form $\Problem$ where questions are mainly about activities/events
happening in a short snippet (TGIF-QA, MSVD-QA, MSRVTT-QA) as well
as long-form $\Problem$ where answers are located at either visual
cues or textual cues as movie subtitles (TVQA dataset). $\ModelName$
demonstrates its competitive reasoning capability on a wide range
of different settings against state-of-the-art methods. The examination
of $\UnitName$ in $\Problem$ highlights the importance of building
a generic neural reasoning unit that supports native multimodal interaction
in improving robustness of visual reasoning.

{\small{}\bibliographystyle{IEEEtran}
\bibliography{thaole}
}{\small\par}
\end{document}